\documentclass[11pt]{article}
\usepackage{latexsym,amssymb,amsmath,amsthm,graphics,graphicx,float,psfrag, epsfig, color, authblk}

\usepackage{xcolor}

\newtheorem{definition}{Definition}

\newtheorem{theorem}{Theorem}
\newtheorem{lemma}{Lemma}

\newcommand{\eps}{\varepsilon}
\newcommand{\R}{\mathbb{R}}

\newcommand{\<}{\langle}
\renewcommand{\>}{\rangle}

\newcommand{\erdosrenyi}{Erd\H{o}s-R\'{e}nyi\ }

\newcommand{\Id}{I_{d\times d}}

\newcommand{\ti}{t_{i}}
\newcommand{\tj}{t_{j}}
\newcommand{\tnot}{t^{(0)}}
\newcommand{\tbar}{\bar{t}}
\newcommand{\tnotbar}{\bar{t}^{(0)}}
\newcommand{\toi}{t^{(0)}_{i}}

\newcommand{\toj}{t^{(0)}_{j}}
\newcommand{\tok}{t^{(0)}_{k}}

\newcommand{\vij}{v_{ij}}

\newcommand{\Eg}{E_g}
\newcommand{\Eb}{E_b}
\newcommand{\proj}{P}
\newcommand{\Pvijp}{\proj_{\vij^\perp}}

\renewcommand{\P}{\operatorname{\mathbb{P}}}
\newcommand{\tonum}[1]{t^{(0)}_{#1}}

\DeclareMathOperator{\Span}{span}

\newcommand{\ttildeoi}{\tilde{t}^{(0)}_i}
\newcommand{\ttildeoj}{\tilde{t}^{(0)}_j}
\newcommand{\vtildeij}{\tilde{v}_{ij}}
\newcommand{\Tbar}{\tbar}
\newcommand{\Tnot}{T^{(0)}}
\newcommand{\Tnotbar}{\overline{t}^{(0)}}

\newcommand{\zij}{z_{ij}}
\newcommand{\tij}{t_{ij}}
\newcommand{\deltaij}{\delta_{ij}}
\newcommand{\deltatilde}{\tilde{\delta}}
\newcommand{\etaij}{\eta_{ij}}
\newcommand{\sij}{s_{ij}}
\newcommand{\lij}{\ell_{ij}}
\newcommand{\loij}{\ell^{(0)}_{ij}}
\newcommand{\toij}{t^{{(0)}}_{ij}}
\newcommand{\tokl}{t^{{(0)}}_{kl}}

\newcommand{\toijperp}{t^{{(0)}\perp}_{ij}}
\newcommand{\toijhat}{\hat{t}^{(0)}_{ij}}
\newcommand{\toikhat}{\hat{t}^{(0)}_{ik}}
\newcommand{\tojkhat}{\hat{t}^{(0)}_{jk}}

\newcommand{\muinf}{\mu_{\infty}}

\title{ShapeFit: Exact location recovery from  corrupted pairwise directions}

\author{ Paul Hand*, Choongbum Lee and Vladislav Voroninski\\
  Department of Mathematics, Massachusetts Institute of Technology\\
  *Department of Computational and Applied Mathematics, Rice University
}

\date{June 4, 2015.  Revised July 4, 2015}		

\begin{document}
\maketitle

\begin{abstract}
Let $t_1,\ldots,t_n \in \mathbb{R}^d$ and consider the location recovery problem: given a subset of pairwise direction observations $\{(t_i - t_j) / \|t_i - t_j\|_2\}_{i<j \in [n] \times [n]}$, where a constant fraction of these observations are arbitrarily corrupted, find $\{t_i\}_{i=1}^n$ up to a global translation and scale. We propose a novel algorithm for the location recovery problem, which consists of a simple convex program over $dn$ real variables. We prove that this program recovers a set of $n$ i.i.d. Gaussian locations exactly and with high probability if the observations are given by an \erdosrenyi graph, $d$ is large enough, and provided that at most a constant fraction of observations involving any particular location are adversarially corrupted.  We also prove that the program exactly recovers  Gaussian locations for $d=3$ if the fraction of corrupted observations at each location is, up to poly-logarithmic factors, at most a constant.  Both of these recovery theorems are based on a set of deterministic conditions that we prove are sufficient for exact recovery.



\end{abstract}

\section{Introduction}

Let $T$ be a collection of $n$ distinct vectors $\tonum{1},\tonum{2},\ldots,\tonum{n} \in \mathbb{R}^d$, and let $G = ([n],E)$ be a graph, where $[n] = \{1,2\ldots,n\}$, and $E = \Eg \sqcup \Eb$, with $\Eb$ and $\Eg$ corresponding to pairwise direction observations that are respectively corrupted and uncorrupted. That is, for each $ij \in E$, we are given a vector $\vij$, where
\begin{align}
\vij = \frac{\toi - \toj}{\bigl \|\toi - \toj \bigr\|_2} \text{ for } ij \in \Eg, \qquad \vij \in \mathbb{S}^{d-1}  \text{ for } ij \in \Eb. \label{measurements}
\end{align}
Thus, an uncorrupted observation $\vij$ is exactly the direction of $\toi$ relative to $\toj$, and a corrupted observation is an arbitrary unit vector. Consider the task of recovering the locations $T$ up to a global translation and scale, from only the observations $\{ v_{ij} \}_{ij \in E}$, and without any knowledge about the decomposition $E = E_g  \sqcup E_b $, nor the nature of the pairwise direction corruptions.

A special case of this problem, with $d=3$, is a necessary subtask in the Structure from Motion (SfM) pipeline for 3D structure recovery from a collection of images taken from different vantage points, a vital aspect of modern computer vision. In the SfM problem, camera locations and orientations are represented as vectors and rotations in $\mathbb{R}^3$, with respect to some global reference frame. Given a collection of images, and for any point in $\R^3$, there is a unique perspective projection of it onto each imaging plane. 
By building local coordinate frames around salient points in the given images, based entirely on photometric information, and comparing them across images, one obtains an estimate of a set of point correspondences. That is, one obtains a set of equivalence classes, where each class corresponds to a physical point in 3D space. Given sufficiently many such sets of point correspondences, epipolar geometry and physical constraints yield estimates of the relative directions and orientations between pairs of cameras. Noise in these estimates is inherent to any real-world application, and worse yet, due to intrinsic challenges arising from the image formation process and properties of man-made scenes (illumination changes, specularities, occlusions, shadows, duplicate structures etc), severe outliers in estimated point correspondences and hence relative camera poses are unavoidable. 


Once camera locations and orientations are estimated, 3D structure can then be recovered by a process called bundle adjustment \cite{1dSfm_22}, which is a simultaneous nonlinear refinement of 3D structure, camera locations, and camera orientations.
Bundle-adjustment is a local method, which generally works well when started close to an optimum. Thus, it is critical to obtain accurate camera location and rotation estimates for initialization. SfM therefore consists of three steps: 1) estimating relative camera pose from point correspondences, 2) recovering camera locations and orientations in a global coordinate framework, and 3) bundle adjustment. While the first and third steps have well-founded theories and algorithms, methods for the second step are mostly heuristically motivated. 

Several efficient and stable algorithms exist for estimating global camera orientations \cite{1dSfm_11,1dSfm_7, 1dSfm_3, 1dSfm_18,1dSfm_8,1dSfm_19,1dSfm_13,1dSfm_5,1dSfm_10, 1dSfm_5,AMIT_13,1dSfm_17,AMIT_25}.  Hence, it is standard to recover locations separately based on estimates of the orientations.






There have been many different approaches to location recovery from relative directions, such as least squares \cite{1dSfm_11, 1dSfm_3,1dSfm_4,1dSfm_17}, second-order cone programs and $l_\infty$ methods \cite{1dSfm_16,1dSfm_17,1dSfm_18, AMIT_19,AMIT_27}, spectral methods \cite{1dSfm_4}, similarity transformations for pair alignment \cite{1dSfm_19}, Lie-algebraic averaging \cite{AMIT_13}, markov random fields \cite{1dSfm_6}, and several others \cite{1dSfm_19,AMIT_32,AMIT_25, 1dSfm_14}. Unfortunately, most location recovery algorithms either lack robustness to correspondence errors (which are unavoidable in large unordered datasets), at times produce illegitimate collapsed solutions, or suffer from convergence to local minima, in sum causing large errors in or a complete degradation of, the recovered locations.

There are some recent notable exceptions to the above limitations.
An algorithm called 1dSfM \cite{1dsfm} focuses on removing outliers by examining inconsistencies along one-dimensional projections, before attempting to recover camera locations. 
However, one drawback of this method is that it does not reason about self-consistent outliers, which occur due to repetitive structures, commonly found in man-made scenes. 
Also, \"{O}zye\c{s}\.{i}l  and Singer propose a convex program over $dn + |E|$ variables for location recovery and empirically demonstrate its robustness to outliers \cite{AMIT}.  While both of these methods exhibit favorable empirical performance, they lack theoretical guarantees of robustness to outliers.

In this paper, we propose a novel convex program for location recovery from pairwise direction observations, and prove that this method recovers locations exactly, in the face of adversarial corruptions, and under rather broad technical assumptions. To the best of our knowledge, this is the first theoretical result guaranteeing location recovery in the challenging case of corrupted pairwise direction observations.
We also demonstrate that this method performs well empirically, recovering locations exactly under severe corruptions of relative directions, and is stable to the simultaneous presence of noise on all the observations, as well as a fraction of arbitrary corruptions.

\subsection{Problem formulation}
The location recovery problem is to recover a set of points in $\mathbb{R}^d$ from observations of pairwise directions between those points.  
Since relative direction observations are invariant under a global translation and scaling, one can at best hope to recover the locations $\Tnot = \{ \tonum{1},\ldots, \tonum{n}\}$ up to such a transformation. That is, successful recovery from $\{v_{ij}\}_{(i,j) \in E}$ is finding a set of vectors ${\{\alpha(\toi + w)\}_{i \in [n]}}$ for some $w \in \mathbb{R}^d$ and $\alpha >0$. We will say that two sets of n vectors $T = \{t_1,\ldots,t_n\}$ and $\Tnot$ are equal up to global translation and scale if there exists a vector $w$ and a scalar $\alpha >0 $ such that $t_i = \alpha(\toi + w)$ for all $i \in [n]$.  In this case, we will say that $T$ and $\Tnot$ have the same `shape,' and we will denote this property as $T \sim \Tnot$. The location recovery problem is then stated as:
\begin{alignat}{2}
&\text{Given:} &\quad &G([n],E), \quad \{\vij\}_{ij \in E} \text{\ \  satisfying \eqref{measurements} }\notag\\
&\text{Find:} && T = \{t_1, \ldots, t_n \} \in \mathbb{R}^{d \times n}, \quad \text{such that} \quad T \sim \Tnot 
\end{alignat}

For this problem to be information theoretically well-posed under arbitrary corruptions, the maximum number of corrupted observations affecting any particular location must be at most $\frac{n}{2}$.  Otherwise, suppose that for some location $\toi$, half of its associated observations $v_{ij}$ are consistent with $\toi$ and the other half are corrupted so as to be consistent with some arbitrary alternative location $w$. Distinguishing between $\toi$ and $w$ is then impossible in general. Formally, let $\deg_b(i)$ be the degree of location $i$ in the graph $([n], \Eb)$. Then well-posedness under adversarial corruption requires that $\max_i \deg_b(i) \leq \gamma n$ for some $\gamma<1/2$. 

Beyond the above necessary degree condition on $E_g$ for well-posedness of recovery, we do not assume anything else about the nature of corruptions. That is, we work with adversarially chosen corrupted edges $E_b$ and arbitrary corruptions of observations associated to those edges. To solve the location recovery problem in this challenging setting, we introduce a simple convex program called ShapeFit:
\begin{align}
\min_{\{t_i\} \in \mathbb{R}^d, i \in [n]} \sum_{ij \in E} \| \Pvijp (t_i - t_j) \|_2 \quad \text{ subject to } \quad \sum_{ij \in E} \langle t_i - t_j, \vij \rangle = 1, \ \  \sum_{i=1}^n t_i = 0 \label{shapefit}
\end{align}
where $\Pvijp$ is the projector onto the orthogonal complement of the span of $\vij$.  

This convex program is a second order cone problem with $dn$ variables and two constraints.  Hence, the search space has dimension $dn-2$, which is minimal due to the $dn$ degrees of freedom in the locations $\{t_i\}$ and the two inherent degeneracies of translation and scale.

\subsection{Main results}

In this paper, we consider the model where pairwise direction observations about $n$  i.i.d. Gaussian locations are given according to an \erdosrenyi random graph.  We start by showing that in a high-dimensional setting, ShapeFit \emph{exactly} recovers the locations with high probability, provided that there are fewer than an exponential number of locations, and provided that at most a fixed fraction of observations are \emph{adversarially} corrupted.

\begin{theorem} \label{thm-complete}
Let $G([n],E)$ be drawn from $G(n,p)$ for some $p = \Omega(n^{-1/4})$. Take $ \tnot_1, \ldots \tnot_n \sim \mathcal{N}(0, I_{d \times d})$ to be i.i.d., independent from $G$. There exists an absolute constant $c>0$ and a $\gamma = \Omega(p^4)$ not depending on $n$, such that if $\max(\frac{2^6}{c^6}, \frac{4^3}{c^3} \log^3 n) \leq  n \leq e^{\frac{1}{6}c d}$ and $d = \Omega(1)$, then there exists an event with probability at least  $1- e^{-n^{1/6}} - 13 e^{-\frac{1}{2}c d}$, on which the following holds:\\[1em]
For arbitrary  subgraphs $\Eb$ satisfying $\max_i \deg_b(i) \leq \gamma n$ and arbitrary pairwise direction corruptions  $\vij \in \mathbb{S}^{d-1}$ for $ij \in \Eb$,  the convex program \eqref{shapefit} has a unique minimizer equal to $\left \{\alpha \Bigl(\toi - \tnotbar \Bigr)\right\}_{i \in [n]}$ for some positive $\alpha$ and for $\tnotbar = \frac{1}{n}\sum_{i\in [n]} \toi$. 
\end{theorem}

This probabilistic recovery theorem is based on a set of deterministic conditions that we prove are sufficient to guarantee exact recovery.  These conditions are satisfied with high probability in the model described above.  See Section \ref{sec-deterministic-theorem} for the deterministic conditions.

This recovery theorem is high-dimensional in the sense that the probability estimate and the exponential upper bound on $n$ are only meaningful for $d= \Omega(1)$.  Concentration of measure  in high dimensions and the upper bound on $n$ ensure control over the angles and distances between random points. As a result, lower dimensional spaces are a more challenging regime for recovery.  

Our other main result is in the physically relevant setting of three-dimensional Euclidean space, where for instance the locations correspond to camera locations.  In this setting, we prove that exact recovery holds for any sufficiently large number of locations, provided that a poly-logarithmically small fraction of observations are adversarially corrupted.

\begin{theorem} \label{thm-3d} 
There exists $n_0 \in \mathbb{N}$ and $c \in \mathbb{R}$ such that the following holds for all $n \ge n_0$.
Let $G([n],E)$ be drawn from $G(n,p)$ for some $p = \Omega( n^{-1/5} \log^{3/5} n)$. Take $ \tnot_1, \ldots \tnot_n  \in \R^3$, where $\toi \sim \mathcal{N}(0, I_{3 \times 3})$ are i.i.d., independent from $G$. 
There exists $\gamma = \Omega(p^5 / \log^3 n)$ and an event of probability at least  $1- \frac{1}{n^{4}}$ 
on which the following holds:\\[1em]
For arbitrary  subgraphs $\Eb$ satisfying $\max_i \deg_b(i) \leq \gamma n$ and arbitrary pairwise direction corruptions $\vij \in \mathbb{S}^2$ for $ij \in \Eb$,  the convex program \eqref{shapefit} has a unique minimizer equal to $\left \{\alpha \Bigl(\toi - \tnotbar \Bigr)\right\}_{i \in [n]}$ for some positive $\alpha$ and for $\tnotbar = \frac{1}{n}\sum_{i\in [n]} \toi$. 
\end{theorem}

Numerical simulations empirically verify the main message of these recovery theorems: ShapeFit recovers a set of locations exactly from corrupted direction observations, provided that up to a constant fraction of the observations at each location are corrupted. We present numerical studies in the setting of locations in $\mathbb{R}^3$, with an underlying random \erdosrenyi graph of observations. Further numerical simulations show that recovery is stable to the additional presence of noise on the uncorrupted measurements.  That is, locations are recovered approximately under such conditions, with a favorable dependence of the estimation error on the measurement noise. 

\



\subsection{Intuition.} \label{sec-motivation}

ShapeFit is a convex program that seeks a set of points whose pairwise directions agree with as many of the corresponding observations as possible.  
The objective, $\sum_{ij \in E} \| \Pvijp (t_i - t_j) \|_2$, incentivizes the correct shape, while permitting translation and a possibly-negative global scale.  Each term $\|\Pvijp (\ti - \tj)\|_2$ is a length-scaled notion for how rotated $t_i - t_j$ is relative to $\pm \vij$.  The objective is in this sense a measure of how much total rotation is needed to deform all $\{t_i - t_j\}_{ij \in E}$ into the observed directions of $\{\pm \vij\}$.  Successful recovery would mean that $\{\|\Pvijp (t_i - t_j)\|_2\}_{ij \in E}$ is sparse.  Motivated by the sparsity promoting properties of $\ell_1$-minimization, the objective in ShapeFit is precisely the $\ell_1$ norm over the edges $E(G)$ of these $\ell_2$ lengths.  

The first constraint in ShapeFit, $\sum_{ij \in E} \langle t_i - t_j, \vij \rangle = 1$, requires that the recovered locations correlate with the provided observations by a strictly positive amount.  It prevents the trivial solution and resolves the global scale ambiguity.  As opposed to the objective, this constraint forbids negative scalings of $\{\toi\}_{i \in [n]}$.  The second constraint, $\sum_{i=1}^n t_i = 0$,  resolves the global translation ambiguity.





\subsection{Organization of the paper}
Section \ref{sec-notation} presents the notation used throughout the rest of the paper.   Section \ref{sec-proofs-high-d} presents the proof of Theorem \ref{thm-complete}.  Section \ref{sec-proofs-three-d} presents the proof of Theorem \ref{thm-3d}. Section \ref{sec-simulations} presents results from numerical simulations.

\subsection{Notation} \label{sec-notation}
Let $[n] = \{1, \ldots, n\}$. Let $e_i$ be the $i$th standard basis element.  
Let $K_n$ be the complete graph on $n$ vertices.  Let $E(K_n)$ be the set of edges in $K_n$. 
Let  $\| \cdot \|_2$ be the standard $\ell_2$ norm on a vector. For any nonzero vector $v$, let $\hat{v} = v / \|v\|_2$.
For a subspace $W$, let $\proj_W$ be the orthogonal projector onto $W$. For a vector $v$, let $\proj_{v^\perp}$ be the orthogonal projector onto the orthogonal complement of the span of $\{v\}$.

Let $T$ denote the set $T = \{ \ti\}_{i \in [n]}$, for $\ti \in \R^{d}$.  Define $\tij = \ti - \tj$
for all distinct $i,j \in [n]$.    
We define $\muinf = \max_{i\neq j} \|\toij\|_2$, and we define $\mu=\frac{1}{|E(G)|}\sum_{ij\in E(G)}\|\toij\|_2$.
Define $\Tbar = \frac{1}{n} \sum_{i \in [n]} \ti$.  Define $\toij$, $\Tnot$, and $\Tnotbar$ similarly.  For a scalar $c$, let $c T = \{c t_i\}_{i\in [n]}$. 
 For a given $G=G([n],E)$ and $\{\vij\}_{ij \in E}$, where $\vij \in \R^{d}$ have unit norm, 
 let $R(T) = \sum_{ij \in E} \| \Pvijp (\ti - \tj) \|_2$.   Let $L(T) = \sum_{ij \in E} \< \ti - \tj, \vij \>$.  Let $\lij = \< \ti - \tj, \vij \>$, and similarly for $\loij$.  In this notation, ShapeFit is
\[
\min_T R(T) \quad \text{subject to} \quad  L(T) = 1, \quad \Tbar = 0
\]
For vectors $v_1, \ldots, v_k$, let $S(v_1, \ldots, v_k) = \Span(v_1, \ldots, v_k)$ be the vector space spanned by these vectors.  
Given $\tij$ and $\toij$, define $\deltaij$, $\etaij$, and $\sij$ such that 
\[
\tij = (1 + \deltaij) \toij + \etaij \sij
\]
where $\sij$ is a unit vector orthogonal to $\toij$ and $\etaij = \| P_{\toijperp} \tij \|_2$.
Note that $\eta_{ij} \ge 0$.

\section{Proof of high dimensional recovery} \label{sec-proofs-high-d}

The proof of Theorem \ref{thm-complete} can be separated into two parts:  a recovery guarantee under a set of deterministic conditions, and a proof that the random model meets these conditions with high probability.  These sufficient deterministic conditions, roughly speaking, are (1) that the graph is connected and the nodes have tightly controlled degrees; (2) that the angles between pairs of locations is uniformly bounded away from $0$ and $\pi$; (3) that all pairwise distances are within a constant factor of each other; (4) that there are not too many corruptions affecting any single location; and (5) that the locations are `well-distributed' relative to each other in a sense we will make precise.   Theorem \ref{thm:mainthm} in Section \ref{sec-deterministic-theorem} states these deterministic conditions formally.

We will prove the deterministic recovery theorem directly, using several geometric properties concerning how deformations of a set of points induce rotations.  
Note that an infinitesimal rigid rotation of two points $\{t_i, t_j\}$ about their midpoint to ${\{t_i + h_i, t_j + h_j\}}$ is such that $h_i - h_j$ is orthogonal to $t_{ij} = t_i - t_j$.  We will abuse terminology and say that $\|P_{\tij^\perp} (h_i - h_j)\|$ is a measure of the rotation in a finite deformation $\{h_i, h_j\}$, and we say that $\langle h_i - h_j, \ti - \tj \rangle$ is the amount of stretching in that deformation. Using this terminology, the geometric properties we establish are:

\begin{itemize}
\item If a deformation stretches two adjacent sides of a triangle at different rates, then that induces a rotation in some edge 
of the triangle (Lemma \ref{lem:rigidity1}).  
\item  If a deformation stretches two nonadjacent sides of a tetrahedron at different rates, then that induces a rotation in some edge 
of the tetrahedron (Lemma \ref{lem:rigidity2}).
\item If a deformation rotates one edge shared by many triangles, then it induces a rotation over many of those triangles, provided the opposite points of those triangles are `well-distributed' (Lemma \ref{lem:triangles}).
\item A deformation that rotates bad edges, must also rotate good edges (Lemma \ref{lem:badgood}).
\item For any deformation, some fraction of the sum of all rotations must  affect the good edges (Lemma \ref{lem:transference}).
\end{itemize}
By using these geometric properties, we show that all nonzero feasible deformations induce a
large amount of total rotation. Since some fraction of the total rotation must be on the good edges,  the objective must increase. 

In Section \ref{sec-deterministic-theorem}, we present the deterministic recovery theorem.  In Section \ref{sec-unbalanced-motions}, we present and prove Lemmas \ref{lem:rigidity1}--\ref{lem:rigidity2}.  In Section \ref{sec-triangles-inequality}, we present and prove Lemmas \ref{lem:triangles}--\ref{lem:transference}.  In Section \ref{sec-proof-mainthm}, we prove the deterministic recovery theorem.  In Section \ref{sec-gaussians-well-distributed}, we prove that Gaussians satisfy several properties, including well-distributedness, with high probability.  In Section \ref{sec-random-graph}, we prove that \erdosrenyi graphs are connected and have controlled degrees and codegrees with high probability.  Finally, in Section \ref{sec-proof-random-theorem}, we prove Theorem \ref{thm-complete}.

%


\subsection{Deterministic recovery theorem in high dimensions} \label{sec-deterministic-theorem}

To state the deterministic recovery theorem, we need two definitions.

\begin{definition}
We say that a graph $G([n],E)$ is \emph{$p$-typical} if it satisfies the following
properties:
\begin{enumerate}
  \setlength{\itemsep}{1pt} \setlength{\parskip}{0pt}
  \setlength{\parsep}{0pt}
	\item $G$ is connected,
	\item each vertex has degree between $\frac{1}{2}np$ and $2np$, and
	\item each pair of vertices has codegree between $\frac{1}{2}np^{2}$ and
$2np^{2}$, where the codegree of a pair of vertices $i,j$ is defined as $|\{k \in [n]\,:\, ik,jk \in E(G)\}|$.
\end{enumerate}
\end{definition}
Note that if $G$ is $p$-typical, then its number of edges is between $\frac{1}{4}n^{2}p$ and $n^{2}p$.

\begin{definition}
Let $T=\{t_{i}\}_{i\in[n]}\subseteq\mathbb{R}^{d}$ be a set of $n$
vectors. Let $G$ be a graph with vertex set $[n]$.
\begin{itemize}
  \setlength{\itemsep}{1pt} \setlength{\parskip}{0pt}
  \setlength{\parsep}{0pt}
\item[(i)] For a pair of vectors $x,y\in\mathbb{R}^{d}$ and a positive real
number $c$, we say that $T$ is \emph{c-well-distributed with respect
to $(x,y)$} if the following holds for all $h\in\mathbb{R}^{d}$:
\[
	\sum_{t \in T}\|\proj_{\Span\{t-x,t-y\}^{\perp}}(h)\|_2 \ge c|T|\cdot\|\proj_{(x-y)^{\perp}}(h)\|_2.
\]
\item[(ii)] We say that $T$ is \emph{$c$-well-distributed along $G$} if
for all distinct $i,j\in[n]$, the set $S_{ij}=\{t_{k}\,:\, ik,jk\in E(G)\}$
is $c$-well-distributed with respect to $(t_{i},t_{j})$.
\end{itemize}
\end{definition}

We now state sufficient deterministic recovery conditions on the graph $G$, the subgraph $\Eb$ corresponding to corrupted observations, and the locations $\Tnot$.

\begin{theorem} \label{thm:mainthm}
Suppose $\Tnot, \Eb, G$ satisfy the conditions
\begin{enumerate}
  \setlength{\itemsep}{1pt} \setlength{\parskip}{0pt}
  \setlength{\parsep}{0pt}
	\item The underlying graph $G$ is $p$-typical,
	\item For all distinct $i,j,k \in [n]$, we have $\sqrt{1-\langle\hat{t}_{ij}^{(0)},\hat{t}_{ik}^{(0)}\rangle^{2}}\ge\beta$,
	\item For all $i,j,k,\ell$ with $i\neq j$ and $k\neq l$, we have $c_{0}\|t_{k\ell}^{(0)}\|_2\le\|t_{ij}^{(0)}\|_2$,
	\item Each vertex has at most $\varepsilon n$ edges in $\Eb$ incident to it,
	\item The set $\{t_{i}^{(0)}\}_{i \in [n]}$ is $c_{1}$-well-distributed
along $G$,
	\item All vectors $\toi$ are distinct,
\end{enumerate}
for constants $0< p, \beta, c_{0}, \eps, c_{1} \leq 1$.  If $\eps \leq \frac{\beta c_0 c_1^2 p^4}{3\cdot 256 \cdot 64 \cdot 32}$, then $L(\Tnot)\neq 0$ and $\Tnot / L(\Tnot)$ is the unique optimizer of ShapeFit.
\end{theorem}

Note that Condition 3 implies that for $\muinf = \max_{i\neq j}\|t_{ij}^{(0)}\|_2$, we have $c_{0}\muinf\le\|t_{ij}^{(0)}\|_2\le\muinf$ for all distinct $i,j \in [n]$. Also note that Conditions 1--6 are invariant under translation and non-zero scalings of $\Tnot$.

Before we prove the theorem, we establish that $L(\Tnot)\neq 0$ when $\eps$ is small enough.  This property guarantees that some scaling of $\Tnot$ is feasible and occurs, roughly speaking, when $|\Eb| < |\Eg|$.

\begin{lemma} \label{lem:T0-feasible} 
If $\varepsilon < \frac{c_0 p}{8}$, then $L(\Tnot) \neq 0$.
\end{lemma}
\begin{proof}
Since $v_{ij} = t_{ij}^{(0)}$ for all $ij \in E_g$, we have
\[
L(T) = \sum_{ij \in E(G)} \langle \toij, v_{ij} \rangle \geq \sum_{ij \in E_g} \| \toij\|_2 - \sum_{ij \in E_b} \| \toij \|_2.
\]
By Condition 3,  $c_0 \muinf |E_g| \leq \sum_{ij \in E_g} \| \toij\|_2$ and $\muinf |E_b| \geq  \sum_{ij\in E_b} \| \toij\|_2$.  Thus it suffices to prove that $c_0 |E_g| > |E_b|$. 
As $\varepsilon < \frac{p}{8}$, Condition 1 and 4 gives $|E_g| \geq \frac{1}{4}n^2 p - \varepsilon n^2 \geq \frac{1}{8}n^2 p$. Since $|E_b| \leq \varepsilon n^2$, if $\varepsilon < \frac{c_0 p}{8}$, then we have $c_0 |E_g| > |E_b|$. 
\end{proof}

The proof of Theorem \ref{thm:mainthm} appears in Section \ref{sec-proof-mainthm}.


\subsection{Unbalanced parallel motions induce rotation} \label{sec-unbalanced-motions}

\begin{lemma} \label{lem:rigidity1}
Let $d\ge2$. Let $t_{1},t_{2},t_{3} \in \R^d$ be distinct.  Let $v_{1},v_{2},v_{3}\in\mathbb{R}^{d}$ and $\alpha \in \R$.   Let $\{\deltatilde_{ij}\}$ be such that  ${\langle v_{i}-v_{j}-\alpha t_{ij},\hat{t}_{ij}\rangle=}$ $\deltatilde_{ij}\|t_{ij}\|_2$ for each distinct $i,j\in[3]$. Then
\[
	\sum_{\substack{i,j\in[3]\\i<j}} 
	\|\proj_{t_{ij}^{\perp}}(v_{i}-v_{j})\|_2
	\ge \sqrt{1-\langle \hat{t}_{12},\hat{t}_{23}\rangle^{2}}\|t_{12}\|_2\left|\deltatilde_{12}-\deltatilde_{13}\right|.
\]
\end{lemma}
\begin{proof}
Note that $t_{ij}=-t_{ji}$ and $\deltatilde_{ij}=\deltatilde_{ji}$ for each
distinct $i,j\in[3]$. Define $W=\Span(\hat{t}_{12},\hat{t}_{23},\hat{t}_{31})$
and define $w_{i}=\proj_{W}v_{i}$ for each $i$. Note that
\[
\sum_{i<j}\|\proj_{t_{ij}^{\perp}}(v_{i}-v_{j})\|_2\ge\sum_{i<j}\|\proj_{t_{ij}^{\perp}}(w_{i}-w_{j})\|_2.
\]
The given condition implies $\proj_{t_{ij}^{\perp}}(w_{i}-w_{j})=w_{i}-w_{j}-(\alpha+\deltatilde_{ij})t_{ij}$
for each distinct $i,j \in [3]$. Therefore,
\begin{eqnarray*}
\sum_{i<j}\|\proj_{t_{ij}^{\perp}}(w_{i}-w_{j})\|_2 & = & \sum_{i<j}\left\| w_{i}-w_{j}-\left(\alpha+\deltatilde_{ij}\right)t_{ij}\right\|_2 \\
 & \ge & \left\| \sum_{(i,j)=(1,2),(2,3),(3,1)}w_{i}-w_{j}-\left(\alpha+\deltatilde_{ij}\right)t_{ij}\right\|_2 \\
 & = & \|\deltatilde_{12}t_{12}+\deltatilde_{23}t_{23}+\deltatilde_{31}t_{31}\|_2.
\end{eqnarray*}
Since $\deltatilde_{13}(t_{12}+t_{23}+t_{31})=0$, the right-hand-side
above equals $\|(\deltatilde_{12}-\deltatilde_{13})t_{12}+(\deltatilde_{23}-\deltatilde_{13})t_{23}\|_2$.
Furthermore,
\begin{eqnarray*}
\left\| (\deltatilde_{12}-\deltatilde_{13})t_{12}+(\deltatilde_{23}-\deltatilde_{13})t_{23}\right\|_2  & \ge & \min_{s\in\mathbb{R}}\|(\deltatilde_{12}-\deltatilde_{13})t_{12}-st_{23}\|_2\\
 & = & \left\| \proj_{t_{23}^{\perp}}(\deltatilde_{12}-\deltatilde_{13})t_{12}\right\|_2 \\
 & \ge & \left|\deltatilde_{12}-\deltatilde_{13}\right|\|t_{12}\|_2\sqrt{1-\langle 
 \hat{t}_{12},\hat{t}_{23}\rangle^{2}}.\qedhere
\end{eqnarray*}
\end{proof}

The previous lemma is applicable only when two
disproportionally scaled edges are incident to each other. The following
lemma shows how to apply the lemma above to the case when we have
two vertex-disjoint edges that are disproportionally scaled.
\begin{lemma} \label{lem:rigidity2}
Let $d\ge2$. Let $t_{1},t_{2},t_{3},t_{4} \in \R^d$ be distinct.  Let $v_{1},v_{2},v_{3},v_{4}\in\mathbb{R}^{d}$ and  $\alpha \in \R$.  Let $\{\deltatilde_{ij}\}$ be such that $\langle v_{i}-v_{j}-\alpha t_{ij},\hat{t}_{ij}\rangle=\deltatilde_{ij}\|t_{ij}\|_2$
for each distinct $i,j\in[4]$. 
Define $\beta=\min\sqrt{1-\langle\hat{t}_{ij},\hat{t}_{ik}\rangle^{2}}$
where the minimum is taken over all distinct $i,j,k\in[4]$ except for the cases when $\{j,k\}=\{1,2\}$. Then
\[
	\sum_{\substack{i,j\in[4]\\i<j}} \|\proj_{t_{ij}^{\perp}}(v_{i}-v_{j})\|_2
	\ge \frac{\beta}{4}\|t_{12}\|_2\left|\deltatilde_{12}-\deltatilde_{34}\right|.
\]
\end{lemma}
\begin{proof}
Note that $t_{ij}=-t_{ji}$ and $\deltatilde_{ij}=\deltatilde_{ji}$ for each
distinct $i,j\in[4]$. Since the given conditions are symmetric under
re-labelling of ($1$ and $2$), and of ($3$ and $4$), we may re-label
if necessary and assume that $\|t_{13}\|_2 \ge\max\{\|t_{14}\|_2 ,\|t_{23}\|_2 ,\|t_{24}\|_2 \}$.
By the triangle inequality, we have $2\|t_{13}\|_2 \ge\|t_{13}\|_2 +\|t_{23}\|_2 \ge\|t_{12}\|_2 $.
Apply Lemma \ref{lem:rigidity1} to the triangle $\{1,2,3\}$ to obtain
\begin{eqnarray}
\sum_{i<j,\ i,j \in \{1, 2, 3\}}\|\proj_{t_{ij}^{\perp}}(v_{i}-v_{j})\|_2 & \ge & \sqrt{1-\langle \hat{t}_{12},\hat{t}_{23}\rangle^{2}}\|t_{12}\|_2 \left|\deltatilde_{12}-\deltatilde_{13}\right|\nonumber \\
 & \ge & \beta\|t_{12}\|_2 \left|\deltatilde_{12}-\deltatilde_{13}\right|,\label{eq:k4_1}
\end{eqnarray}
and similarly apply the lemma to the triangle $\{3,1,4\}$ to obtain
\begin{eqnarray}
\sum_{i<j, \ i,j\in\{1, 3, 4\}}\|\proj_{t_{ij}^{\perp}}(v_{i}-v_{j})\|_2 & \ge & \sqrt{1-\langle \hat{t}_{13},\hat{t}_{14}\rangle^{2}}\|t_{13}\|_2 \left|\deltatilde_{13}-\deltatilde_{34}\right|\nonumber \\
 & \ge & \beta\|t_{13}\|_2 |\deltatilde_{13}-\deltatilde_{34}|\ge\frac{\beta}{2}\|t_{12}\|_2 |\deltatilde_{13}-\deltatilde_{34}|.\label{eq:k4_2}
\end{eqnarray}
By adding \eqref{eq:k4_1} and \eqref{eq:k4_2}, we see that
\begin{align*}
	&\,\sum_{i<j,\ i,j \in \{1, 2, 3\}}\|\proj_{t_{ij}^{\perp}}(v_{i}-v_{j})\|_2
	+
	\sum_{i<j, \ i,j\in\{1, 3, 4\}}\|\proj_{t_{ij}^{\perp}}(v_{i}-v_{j})\|_2 \\
	&\,\ge \,\, 
	\beta\|t_{12}\|_2 \left|\deltatilde_{12}-\deltatilde_{13}\right|
	+
	\frac{\beta}{2}\|t_{12}\|_2 \left|\deltatilde_{13}-\deltatilde_{34}\right|
	\ge
	\frac{\beta}{2}\|t_{12}\|_2 \left|\deltatilde_{12}-\deltatilde_{34}\right|.
\end{align*}
The lemma follows since the left-hand-side is bounded from
above by $2 \sum_{\substack{i,j\in[4]\\i<j}} \|\proj_{t_{ij}^{\perp}}(v_{i}-v_{j})\|_2$.
\end{proof}


\subsection{Triangles inequality and rotation propagation} \label{sec-triangles-inequality}

\begin{lemma}[Triangles Inequality] \label{lem:triangles}
Let $d\ge3$; $x,y,t_{1},t_{2},\cdots,t_{k}\in\mathbb{R}^{d}$.
If $T=\{t_{1},\cdots,t_{k}\}$ is $c$-well-distributed
with respect to $(x,y)$, then for all vectors $h_{x},h_{y},h_{1},\cdots,h_{k}\in\mathbb{R}^{d}$
and sets $X\subseteq[k]$, we have
\[
\sum_{i\in[k]\setminus X}\|\proj_{(x-t_{i})^{\perp}}(h_{x}-h_{i})\|_2 +\|\proj_{(t_{i}-y)^{\perp}}(h_{i}-h_{y})\|_2 \ge(ck-|X|)\cdot\|\proj_{(x-y)^{\perp}}(h_{x}-h_{y})\|_2.
\]
\end{lemma}
\begin{proof}
For each $i\in[k]$, define $W_{i}=\Span\langle x-t_{i}\,,\, t_{i}-y\rangle$.
Define $P$ as the projection map to the space of vectors orthogonal
to $x-y$, and define $P_{i}$ for each $i\in[k]$  as the projection
map to $W_{i}^{\perp}$. Since $(x-t_{i})^{\perp}\supseteq W_{i}^{\perp}$
and $(t_{i}-y)^{\perp}\supseteq W_{i}^{\perp}$, it follows that 
\begin{eqnarray*}
 &  & \sum_{i\in[k]\setminus X}\|\proj_{(x-t_{i})^{\perp}}(h_{x}-h_{i})\|_2 +\|\proj_{(t_{i}-y)^{\perp}}(h_{i}-h_{y})\|_2\\
 & \ge & \sum_{i\in[k]\setminus X}\|P_{i}(h_{x}-h_{i})\|_2 +\|P_{i}(h_{i}-h_{y})\|_2 \ge\sum_{i\in[k]\setminus X}\|P_{i}(h_{x}-h_{y})\|_2.
\end{eqnarray*}
Since $t_{1},\cdots,t_{k}$ are well-distributed with respect to $(x,y)$,
we have
\begin{equation}
\sum_{i\in[k]}\|P_{i}(h_{x}-h_{y})\|_2 \ge ck\cdot\|P(h_{x}-h_{y})\|_2.\label{eq:proj_triangles}
\end{equation}
Since $\|P_{i}(h_{x}-h_{y})\|_2\le\|P(h_{x}-h_{y})\|_2$ holds for all
$i$, it follows that 
\[
\sum_{i\in[k]\setminus X}\|P_{i}(h_{x}-h_{y})\|_2 \ge(ck-|X|)\cdot\|P(h_{x}-h_{y})\|_2,
\]
proving the lemma.
\end{proof}

The proof of Theorem \ref{thm:mainthm} will rely on the following two lemmas, which state that rotational motions on some parts of the graph bound rotational motions on other parts. The following lemma relates the rotational motions on bad edges to the rotational motions on good edges.  Recall the notation $\tij = (1 + \deltaij) \toij + \etaij \sij$ where $\sij$ is a unit vector orthogonal to $\toij$ and $\etaij = \| P_{\toijperp} \tij \|_2$.

\begin{lemma} \label{lem:badgood} 
Fix $T$. If $\varepsilon_0 \leq \frac{c_1 p^2}{8}$, then $\sum_{ij\in E_{g}}\eta_{ij}\ge\frac{c_{1}p^{2}}{8\varepsilon_0}\sum_{ij\in E_{b}}\eta_{ij}$. 
\end{lemma}
\begin{proof}
For each edge $ij \in E(K_n)$, by Conditions \ref{enu:typical}, \ref{enu:bad}, \ref{enu:welldistributed}; Lemma \ref{lem:triangles} and $\varepsilon_0 \leq \frac{c_1p^2}{8}$,
we have 
\[
	\sum_{\substack{k\neq i,j\\ik,jk\in E_{g}}}(\eta_{ik}+\eta_{jk})
	\ge\left(c_{1}\cdot\frac{1}{2}np^{2}-2\varepsilon_0n\right)\cdot\eta_{ij}
	\ge\frac{c_{1}}{4}np^{2}\cdot\eta_{ij}.
\]
Therefore, if we sum the inequality above for all bad edges $ij\in E_{b}$, then
\[
	\sum_{ij\in E_{b}}\sum_{\substack{k\neq i,j\\ik,jk\in E_{g}}} (\eta_{ik}+\eta_{jk})
	\ge\frac{c_{1}}{4}np^{2}\cdot\sum_{ij\in E_{b}}\eta_{ij}.
\]
For fixed $ik\in E_{g}$, the left-hand-side may sum $\eta_{ik}$
as many times as the number of bad edges incident to the edge $ik$.
Hence by Condition 4, the left-hand-side of above is at most 
\[
	\sum_{ij\in E_{b}} \sum_{\substack{k\neq i,j\\ik,jk\in E_{g}}} (\eta_{ik}+\eta_{jk})
	\le 2\varepsilon_0 n\cdot\sum_{ij\in E_{g}}\eta_{ij}.
\]
Therefore by combining the two inequalities above, we obtain 
\[
	\sum_{ij\in E_{b}}\eta_{ij}
	\le\frac{8\varepsilon_0}{c_{1}p^{2}}\sum_{ij\in E_{g}}\eta_{ij}. 
	\qedhere
\]
\end{proof}

The following lemma relates the rotational motions over the good graph $\Eg$ to rotational motions over the complete graph $K_n$.

\begin{lemma} \label{lem:transference} 
Fix $T$. If $\varepsilon_0 \leq \frac{c_1 p^2}{8}$, then $\sum_{ij\in E_{g}}\eta_{ij}\ge\frac{c_{1}p}{16}\sum_{ij\in E(K_{n})}\eta_{ij}$.
\end{lemma}
\begin{proof}
For each $ij\in E(K_{n})$, since $\{t_{i}^{(0)}\}_{i=1}^{n}$ is
$c_{1}$-well-distributed along $G$ and $G$ is $p$-typical, we
have as in the proof of Lemma~\ref{lem:badgood},
\[
	\sum_{\substack{k\neq i,j\\ik,jk\in E_{g}}}(\eta_{ik}+\eta_{jk})
	\ge \left(c_{1}\cdot\frac{1}{2}np^{2}-2\varepsilon_0 n\right) \cdot\eta_{ij} 
	\ge \frac{c_{1}}{4}np^{2}\cdot\eta_{ij}.
\]
If we sum the above over all $ij\in E(K_{n})$, we obtain 
\[
	\sum_{ij\in E(K_{n})}\sum_{\substack{k\neq i,j\\ik,jk\in E_{g}}}(\eta_{ik}+\eta_{jk})
	\ge \frac{c_{1}}{4}np^{2} \cdot\sum_{ij\in E(K_{n})}\eta_{ij}.
\]
For a fixed $ik\in E_{g}$, the left-hand-side may sum $\eta_{ik}$
as many as times as the number of edges of $G$ incident to $ik$. Therefore
since $G$ is $p$-typical, we see that 
\[
	\sum_{\substack{k\neq i,j\\ik,jk\in E_{g}}}(\eta_{ik}+\eta_{jk})
	\le 2\cdot2np\sum_{ij\in E_{g}}\eta_{ij}.
\]
By combining the two inequalities, we obtain 
\[
 	\frac{c_{1}}{4}np^{2}\sum_{ij\in E(K_{n})}\eta_{ij}
 	\le 4np\sum_{ij\in E_{g}}\eta_{ij},
\]
and thus $\frac{c_{1}p}{16}\sum_{ij\in E(K_{n})}\eta_{ij}\le\sum_{ij\in E_{g}}\eta_{ij}.$
\end{proof}

\subsection{Proof of Theorem \ref{thm:mainthm}} \label{sec-proof-mainthm}

We now prove the deterministic recovery theorem.

\begin{proof}[Proof of Theorem \ref{thm:mainthm}]

By Lemma \ref{lem:T0-feasible} and the fact that Conditions 1--6 are invariant under global translation and nonzero scaling, we can take $\Tnotbar=0$ and $L(\Tnot) = 1$ without loss of generality.  The variable $\muinf$ is to be understood accordingly.

We will directly prove that $R(T) > R(\Tnot)$ for all $T\neq \Tnot$ such that $L(T)=1$ and $\Tbar = 0$. Consider an arbitrary feasible $T$ and recall the notation $\tij = (1 + \deltaij) \toij + \etaij \sij$ where $\sij$ is a unit vector orthogonal to $\toij$ and $\etaij = \| P_{\toijperp} \tij \|_2$. A useful lower bound for the objective $R(T)$ is given by 
\begin{eqnarray}
	R(T)
	=\sum_{ij}\|\proj_{v_{ij}^{\perp}}t_{ij}\| _2
	& = & \sum_{ij\in E_{g}}\eta_{ij}+\sum_{ij\in E_{b}}\|\proj_{v_{ij}^{\perp}}t_{ij}\|_2\nonumber \\
	& \ge & \sum_{ij\in E_{g}}\eta_{ij}+\sum_{ij\in E_{b}}\left(\|\proj_{v_{ij}^{\perp}}t_{ij}^{(0)}\|_2-|\delta_{ij}|\|t_{ij}^{(0)}\|_2-\eta_{ij}\right)\nonumber \\
	& \ge & R(\Tnot)+\sum_{ij\in E_{g}}\eta_{ij}-\sum_{ij\in E_{b}}(|\delta_{ij}|\|t_{ij}^{(0)}\|_2+\eta_{ij}).\label{eq:gain3}
\end{eqnarray}

Suppose that $\sum_{ij\in E_{b}}|\delta_{ij}|\|t_{ij}^{(0)}\|_2<\sum_{ij\in E_{b}}\eta_{ij}$.
Since Lemma \ref{lem:badgood} for $\varepsilon\le\frac{c_{1}p^2}{16}$
implies $\sum_{ij\in E_{b}}\eta_{ij}\le\frac{1}{2}\sum_{ij\in E_{g}}\eta_{ij}$,
by (\ref{eq:gain3}), we have
\begin{eqnarray*}
	R(T) & \ge & R(\Tnot)+\sum_{ij\in E_{g}}\eta_{ij}-\sum_{ij\in E_{b}}(|\delta_{ij}|\|t_{ij}^{(0)}\|_2+\eta_{ij})\\
	& > & R(\Tnot)+\sum_{ij\in E_{g}}\eta_{ij}-\sum_{ij\in E_{b}}2\eta_{ij}\ge R(\Tnot).
\end{eqnarray*}
Hence we may assume 
\begin{equation}
	\sum_{ij\in E_{b}}|\delta_{ij}|\|t_{ij}^{(0)}\|_2\ge\sum_{ij\in E_{b}}\eta_{ij}.\label{eq:badtwist}
\end{equation}

In the case $|\Eb| \neq 0$, define $\overline{\delta}=\frac{1}{|E_{b}|}\sum_{ij\in E_{b}}|\delta_{ij}|$
as the average `relative parallel motion' on the bad edges. For distinct edges $ij,k\ell\in E(K_{n})$,
if $\{i,j\}\cap\{k,\ell\} = \emptyset$, then define $\eta(ij,k\ell)=\eta_{ij}+\eta_{ik}+\eta_{i\ell}+\eta_{jk}+\eta_{j\ell}+\eta_{k\ell}$,
and if $\{i,j\}\cap\{k,\ell\}\neq\emptyset$ (without loss of generality,
assume $\ell=i$), then define $\eta(ij,k\ell)=\eta_{ij}+\eta_{ik}+\eta_{jk}$. \\

\noindent \textbf{Case 0}. $\bar{\delta} = 0$ or $|\Eb| = 0$.
\medskip

Note that $\bar{\delta}=0$ implies $\delta_{ij} =0$ for all $ij \in E_b$, which by \eqref{eq:badtwist} implies
$\eta_{ij} =0$ for all $ij \in E_b$.  Therefore by \eqref{eq:gain3}, we have
\[
	R(T) \geq R(\Tnot) + \sum_{ij \in E_g} \eta_{ij}.
\]
If $\sum_{ij \in E_g} \eta_{ij} > 0$, then we have $R(T) > R(\Tnot)$. Thus 
we may assume that $\eta_{ij} = 0$ for all $ij \in E_g$. In this case, we will show that $T = \Tnot$.

By Lemma~\ref{lem:transference}, if $\varepsilon_0 \le \frac{c_1p^2}{8}$, then
$\eta_{ij}=0$ for all $ij \in E(G)$ implies that
$\eta_{ij} = 0$ for all $ij \in E(K_n)$. 
For $ij \in E_b$, since $\delta_{ij} = \eta_{ij} = 0$, it follows that $\ell_{ij} = \ell_{ij}^{(0)}$.
Since $\delta_{ij} \|\toij\|_2 = \ell_{ij} - \ell_{ij}^{(0)}$ for $ij \in E_g$, we have
\[
	0
	= \sum_{ij \in E(G)} (\ell_{ij} - \ell_{ij}^{(0)})
	= \sum_{ij \in E_b} (\ell_{ij} - \ell_{ij}^{(0)}) + \sum_{ij \in E_g} (\ell_{ij} - \ell_{ij}^{(0)})
	= \sum_{ij \in E_g} (\ell_{ij} - \ell_{ij}^{(0)})
	= \sum_{ij \in E_g} \delta_{ij} \|\toij\|_2,
\] 
where the first equality is because $L(T) = L(\Tnot) = 1$.
By Condition 6, $\|\toij\|_2 \neq 0$ for all $i \neq j$. Therefore, if $\delta_{ij} \neq 0$ for some $ij \in E_g$, then there exists $ab, cd \in E_g$ 
such that $\delta_{ab} > 0$ and $\delta_{cd} < 0$. 
By Lemma~\ref{lem:rigidity1} or \ref{lem:rigidity2} and Condition 2, this forces $\eta(ab,cd) > 0$, contradicting the fact that
$\eta_{ij} = 0$ for all $ij \in E(K_n)$. 
Therefore $\delta_{ij} = 0$ for all $ij \in E_g$, and hence
$\delta_{ij} = 0$ for all $ij \in E(G)$.  

Define $t_i = t_i^{(0)} + h_i$ for each $i \in [n]$. Because $\eta_{ij} = \delta_{ij} = 0$ for all $ij \in E(G)$, we have $h_i = h_j$ 
for all $ij \in E(G)$. Since $G$ is connected, this implies $h_i = h_j$ for all 
$i \in [n]$. Then by the constraint $\sum_{i \in [n]} t_i = \sum_{i \in [n]} t_i^{(0)} = 0$, 
we get $h_i = 0$ for all $i \in [n]$. Therefore $T = \Tnot$. \\

\noindent\textbf{Case 1}. $\bar{\delta} \neq 0$ and $\sum_{ij\in E_{g}}|\delta_{ij}|<\frac{1}{8}\overline{\delta}|E_{g}|$ and $|\Eb| \neq 0$. 
\medskip

Define $L_{b}=\{ij\in E_{b}:|\delta_{ij}|\ge\frac{1}{2}\overline{\delta}\}$.
Note that $\sum_{ij\in E_{b}\setminus L_{b}}|\delta_{ij}|<\frac{1}{2}\overline{\delta}|E_{b}|$
and therefore 
\begin{equation}
\sum_{ij\in L_{b}}|\delta_{ij}|=\sum_{ij\in E_{b}}|\delta_{ij}|-\sum_{ij\in E_{b}\setminus L_{b}}|\delta_{ij}|>\sum_{ij\in E_{b}}|\delta_{ij}|-\frac{1}{2}\overline{\delta}|E_{b}|=\frac{1}{2}\overline{\delta}|E_{b}|.\label{eq:large_delta}
\end{equation}
Define $F_{g}=\{ij\in E_{g}:|\delta_{ij}|<\frac{1}{4}\overline{\delta}\}$.
Then by the condition of Case 1,
\[
\frac{1}{8}\overline{\delta}|E_{g}|>\sum_{ij\in E_{g}}|\delta_{ij}|\ge\sum_{ij\in E_{g}\setminus F_{g}}|\delta_{ij}|\ge\frac{1}{4}\overline{\delta}|E_{g}\setminus F_{g}|,
\]
and therefore $|E_{g}\setminus F_{g}|<\frac{1}{2}|E_{g}|$, or equivalently,
$|F_{g}|>\frac{1}{2}|E_{g}|$.

For each $ij\in L_{b}$ and $k\ell\in F_{g}$, by Lemmas \ref{lem:rigidity1}, \ref{lem:rigidity2},
and Condition 3, we have $\eta(ij,k\ell)\ge \frac{\beta}{4}|\delta_{ij} - \delta_{k\ell}| \cdot\|t_{ij}\|_2 \ge
\frac{\beta}{4} \cdot \frac{1}{2}|\delta_{ij}| \cdot \|t_{ij}\|_2 \ge
\frac{\beta c_{0}\muinf}{8}|\delta_{ij}|$. 
Therefore by Condition 1, 
\begin{eqnarray*}
	\sum_{ij\in E_{b}}\sum_{k\ell\in E_{g}}\eta(ij,k\ell) 
	& \ge & \sum_{ij\in L_{b}}\sum_{k\ell\in F_{g}}\frac{\beta c_{0}\muinf}{8}|\delta_{ij}|=\sum_{ij\in L_{b}}|F_{g}|\cdot\frac{\beta c_{0}\muinf}{8}|\delta_{ij}|\\
	& >  & \sum_{ij\in L_{b}}\frac{\beta c_{0}\muinf}{16}|E_{g}||\delta_{ij}|
	\ge\frac{\beta c_{0}\muinf}{16}|E_{g}|\cdot\frac{1}{2}\overline{\delta}|E_{b}|,
\end{eqnarray*}
where the last inequality follows from (\ref{eq:large_delta}). For
each $ij\in E(K_{n})$, we would like to count how many times each
$\eta_{ij}$ appear on the left hand side. If $ij\in E_{b}$, then
there are at most ${n \choose 2}$ $K_{4}$s and $n$ $K_{3}$s containing
$ij$; hence $\eta_{ij}$ may appear at most $6{n \choose 2}+3n=3n^{2}$
times. If $ij\notin E_{b}$, then $\eta_{ij}$ appears when there is
a $K_{4}$ or a $K_{3}$ containing $ij$ and some bad edge. By Condition
4, there are at most $2\varepsilon n$ such bad $K_{3}$s. If the
bad edge in $K_{4}$ is incident to $ij$, then there are at most
$2\varepsilon n\cdot(n-3)$ such $K_{4}$s, and if the bad edge is
not incident to $ij$, then there are at most $|E_{b}|\le\varepsilon n^{2}$
such $K_{4}$. Thus $\eta_{ij}$ may appear at most $3\cdot2\varepsilon n+6\cdot(2\varepsilon n(n-3)+\varepsilon n^{2})\le18\varepsilon n^{2}$
times. Therefore
\begin{eqnarray*}
	\sum_{ij\in E_{b}}\sum_{k\ell\in E_{g}}\eta(ij,k\ell) 
	& \le & \sum_{ij\in E_{b}}3n^{2}\cdot\eta_{ij}+\sum_{ij\in E(K_{n})}18\varepsilon n^{2}\cdot\eta_{ij}.
\end{eqnarray*}
By Lemma \ref{lem:badgood}, if $\varepsilon < \frac{c_1p^2}{8}$, we have
\[
	\sum_{ij\in E_{b}}\sum_{k\ell\in E_{g}}\eta(ij,k\ell)
	\le\frac{24\varepsilon}{c_{1}p^{2}}n^{2}\sum_{ij\in E_{g}}\eta_{ij}+\sum_{ij\in E(K_{n})}18\varepsilon n^{2}\cdot\eta_{ij}
	\le\frac{42\varepsilon}{c_{1}p^{2}}n^{2}\sum_{ij\in E(K_{n})}\eta_{ij}.
\]
Hence
\[
	\frac{42\varepsilon}{c_{1}p^{2}}n^{2}\sum_{ij\in E(K_{n})}\eta_{ij}\ge\frac{\beta c_{0}\muinf}{32}|E_{g}|\cdot\overline{\delta}|E_{b}|.
\]
If $\varepsilon< \frac{p}{8}$, then $|E_{g}|\ge\frac{n^{2}p}{4}-|E_{b}|\ge\frac{n^{2}p}{8}$. 
Further, if $\varepsilon< \frac{\beta c_0 c_1^2 p^4}{32 \cdot 42 \cdot 32 \cdot 8}$, then by Condition 3,
 $\bar{\delta} \neq 0$, and $|\Eb| \neq 0$, the above implies 
\begin{align*}
	\sum_{ij\in E(K_{n})}\eta_{ij}
	\ge&\, \frac{\beta c_{0}c_{1}p^{2}}{42\cdot 32 \varepsilon n^{2}}\muinf|E_{g}|\cdot\overline{\delta}|E_{b}| 
	\ge \frac{\beta c_{0}c_{1}p^{3}}{42\cdot 32 \cdot 8} \cdot \frac{1}{\varepsilon} \cdot \muinf \overline{\delta}|E_{b}|  \\
	>&\, \frac{32}{c_{1}p}\muinf\cdot\overline{\delta}|E_{b}|
	\ge \frac{32}{c_{1}p}\sum_{ij\in E_{b}}|\delta_{ij}|\|t_{ij}^{(0)}\|_2.
\end{align*}
Lemma \ref{lem:transference} implies 
\[
	\sum_{ij\in E_{g}}\eta_{ij}
	\ge \frac{c_{1}p}{16}\sum_{ij\in E(K_{n})}\eta_{ij}
	> 2\sum_{ij\in E_{b}}|\delta_{ij}|\|t_{ij}^{(0)}\|_2.
\]
Therefore by (\ref{eq:badtwist}),we have $\sum_{ij\in E_{g}}\eta_{ij}>\sum_{ij\in E_{b}}(|\delta_{ij}|\|t_{ij}^{(0)}\|_2+\eta_{ij})$ if $\varepsilon \le \min\{\frac{c_1p^2}{8}, \frac{p}{8}, \frac{\beta c_0 c_1^2 p^4}{32 \cdot 42 \cdot 32 \cdot 8} \}$.
By (\ref{eq:gain3}), this shows $R(T)>R(\Tnot)$. This condition on $\eps$ is satisfied under the assumption $\eps \leq \frac{\beta c_0 c_1^2 p^4}{3\cdot 256 \cdot 64 \cdot 32}$.
\\

\noindent\textbf{Case 2}. $\bar{\delta} \neq 0$ and $\sum_{ij\in E_{g}}|\delta_{ij}|\ge\frac{1}{8}\overline{\delta}|E_{g}|$  and $|\Eb| \neq 0$. 
\medskip

Define $E_{+}=\{ij\in E_{g}\,:\,\delta_{ij}\ge0\}$ and $E_{-}=\{ij\in E_{g}\,:\,\delta_{ij}<0\}$.
Since $\ell_{ij}-\ell_{ij}^{(0)}=\delta_{ij}\|t_{ij}^{(0)}\|_2$
for $ij\in E_{g}$, we have 
\begin{eqnarray*}
0=\sum_{ij \in E(G)}(\ell_{ij}-\ell_{ij}^{(0)}) & = & \sum_{ij\in E_{b}}(\ell_{ij}-\ell_{ij}^{(0)})+\sum_{ij\in E_{g}}\delta_{ij}\|t_{ij}^{(0)}\|_2.
\end{eqnarray*}
where the first equality follows from $L(T) = L(\Tnot)$.  Therefore, 
\begin{eqnarray*}
	\left| \sum_{ij\in E_{g}}\delta_{ij}\|t_{ij}^{(0)}\|_2 \right|
	\le \left| \sum_{ij\in E_{b}}(\ell_{ij}-\ell_{ij}^{(0)}) \right|
	\le \sum_{ij\in E_{b}}(|\delta_{ij}|\|t_{ij}^{(0)}\|_2+\eta_{ij}) 
	\le 2\muinf\overline{\delta}|E_{b}|,
\end{eqnarray*}
where the last inequality follows from (\ref{eq:badtwist}), Condition 3, and the definition of $\overline{\delta}$. On the other hand, the condition of Case 2 and Condition 3 implies 
$\sum_{ij\in E_{g}}|\delta_{ij}|\|t_{ij}^{(0)}\|_2 \ge \frac{1}{8}c_0 \muinf\overline{\delta}|E_g|$. 
Therefore
\[
	\sum_{ij \in E_-} (-\delta_{ij}) \|t_{ij}^{(0)}\|_2
	= \frac{1}{2}\left( -\sum_{ij\in E_{g}}\delta_{ij}\|t_{ij}^{(0)}\|_2 + \sum_{ij\in E_{g}}|\delta_{ij}|\|t_{ij}^{(0)}\|_2\right)
	\ge \frac{1}{2} \left(\frac{1}{8} c_0 \muinf\overline{\delta}|E_g| - 2\muinf\overline{\delta}|E_{b}|\right).
\]
If $\varepsilon \le \frac{1}{256}c_0p$, then 
since $|E_b| \le \varepsilon n^2$ and $|E_g| \ge \frac{1}{4}n^2p - |E_b| \ge \frac{1}{8}n^2p$, 
we see that $\frac{1}{8} c_0 \muinf\overline{\delta}|E_g| - 2\muinf\overline{\delta}|E_{b}| \ge \frac{1}{16}c_0\muinf \bar{\delta}|E_g|$.
Therefore $\sum_{ij\in E_{-}}(-\delta_{ij})\|t_{ij}^{(0)}\|_2 \ge\frac{1}{32}c_{0}\muinf \overline{\delta}|E_{g}|$.
Similarly, $\sum_{ij\in E_{+}} \delta_{ij}\|t_{ij}^{(0)}\|_2 \ge \frac{1}{32}c_{0}\muinf \overline{\delta}|E_{g}|$.

If $|E_+| \ge \frac{1}{2}|E_g|$, then 
by Lemmas \ref{lem:rigidity1}, \ref{lem:rigidity2}, and Condition 3, we have 
\begin{eqnarray*}
	\sum_{ij\in E_{-}}\sum_{k\ell\in E_{+}}\eta(ij,k\ell) 
	& \ge & \sum_{ij\in E_{-}}\sum_{k\ell\in E_{+}}\frac{\beta}{4}(-\delta_{ij})\|t_{ij}^{(0)}\|_2\\
	& \ge & \sum_{ij\in E_{-}}(-\delta_{ij})\|t_{ij}^{(0)}\|_2\cdot \frac{\beta}{4}|E_{+}|
	\ge \frac{\beta}{4}|E_{+}|\cdot\frac{1}{32}c_{0}\muinf \overline{\delta}|E_{g}| \\
	& \ge & \frac{\beta}{256} c_{0} \muinf\overline{\delta}|E_{g}|^2. 
\end{eqnarray*}
Similarly, if $|E_-| \ge \frac{1}{2}|E_g|$, then we can switch the order of summation and
consider $\sum_{ij \in E_+} \sum_{k\ell \in E_-} \eta(ij, k\ell)$ to obtain the 
same conclusion.

Since each edge is contained in at most $\frac{n(n-1)}{2}$
copies of $K_{4}$ and $n$ copies of $K_{3}$ (and there are 6 edges
in a $K_{4}$, 3 edges in a $K_3$), we have 
\[
	\sum_{ij\in E_{-}}\sum_{k\ell\in E_{+}}\eta(ij,k\ell)
	\le\left(6\frac{n(n-1)}{2}+3n\right)\sum_{ij\in E(K_{n})}\eta_{ij}\le3n^{2}\sum_{ij\in E(K_{n})}\eta_{ij}.
\]
If $\varepsilon \le \frac{p}{8}$, then $|E_{g}|\ge \frac{1}{4}n^{2}p-|E_{b}| \ge\frac{1}{8}n^{2}p$.
Further, if $\varepsilon < \frac{\beta c_0 c_1 p^3}{3 \cdot 256 \cdot 64 \cdot 32}$, then since $\bar{\delta} \neq 0$ and $|\Eb| \leq \eps n^2$, we have
\[
	\sum_{ij\in E(K_{n})}\eta_{ij}
	\ge\frac{1}{3n^{2}}\cdot\frac{\beta c_{0} \muinf \overline{\delta}}{256} |E_{g}|^2
	\ge\frac{\beta c_{0} p^2}{3\cdot 256 \cdot 64}\muinf \overline{\delta} n^2
	>\frac{32}{c_{1}p}\muinf\overline{\delta}|E_{b}|.
\]
By Lemma \ref{lem:transference}, if $\varepsilon < \frac{c_1p^2}{8}$, then this implies 
\[
\sum_{ij\in E_{g}}\eta_{ij}\ge\frac{c_{1}p}{16}\sum_{ij\in E(K_{n})}\eta_{ij} > 2\muinf\overline{\delta}|E_{b}|.
\]
Therefore from (\ref{eq:gain3}), (\ref{eq:badtwist}), and Condition 3, 
if $\varepsilon \le \min\{\frac{c_0 p}{256}, \frac{c_1p^2}{8}, \frac{p}{8}, \frac{\beta c_0 c_1 p^3}{3 \cdot 256 \cdot 64 \cdot 32} \} $, then
\begin{eqnarray*}
R(T) & \ge & R(\Tnot)+\sum_{ij\in E_{g}}\eta_{ij}-\sum_{ij\in E_{b}}(|\delta_{ij}|\|t_{ij}^{(0)}\|_2+\eta_{ij})\\
 & > & R(\Tnot)+2\muinf\overline{\delta}|E_{b}|-\sum_{ij\in E_{b}}2|\delta_{ij}|\|t_{ij}^{(0)}\|_2\ge R(\Tnot).
\end{eqnarray*}
This condition on $\eps$ is satisfied under the assumption $\eps \leq \frac{\beta c_0 c_1^2 p^4}{3\cdot 256 \cdot 64 \cdot 32}$.
\end{proof}

\subsection{Properties of Gaussians in high dimensions} \label{sec-gaussians-well-distributed}

In this section, we prove that i.i.d. Gaussians satisfy properties needed to establish Conditions 2, 3, and 5 in Theorem \ref{thm:mainthm}. We begin by recording some useful facts regarding concentration of random Gaussian vectors:

\begin{lemma}
\label{lem:conc1}
Let $x,y$ be i.i.d. $\mathcal{N}(0,I_{d\times d})$, and $\epsilon \leq 1$, then
\[
\P\left ( d(1-\epsilon) \leq \|x\|_2^2 \leq d(1+\epsilon) \right) \geq 1- e^{-c \epsilon^2 d}
\]
and
\[
\P \left( |\langle x, y \rangle| \geq d\epsilon \right) \leq e^{-c\epsilon^2 d}
\]
where $c>0$ is an absolute constant.
\end{lemma}
\begin{proof}
Both statements follow from Corollary 5.17 in \cite{Vershynin}, concerning concentration of sub-exponential random variables.
\end{proof}

\begin{lemma}[\cite{Vershynin} Corollary 5.35]
\label{lem:conc2}
Let $A$ be an $n \times d$ matrix with i.i.d. $\mathcal{N}(0,1)$ entries. Then for any $t\geq 0$,
\[
\P\left( \sigma_{\max}(A) \geq \sqrt{n} + \sqrt{d} + t \right) \leq 2e^{-\frac{t^2}{2}}
\]
where $\sigma_{\max}(A)$ is the largest singular value of $A$.
\end{lemma}


\begin{lemma}
\label{lem:conc3}
Let $\toi, i \in [n]$ be i.i.d. $\mathcal{N}(0,I_{d \times d})$. Then, there exists an event $E$, such that on E, we have for all $i,j,k,l \in [n], i \neq j, k\neq l$,
\[
\frac{\|\toij\|_2}{\|\tokl\|_2} \geq \frac{9}{10}
\]
and for all distinct $i,j,k \in [n]$,
\[
\langle \toijhat,\toikhat \rangle^2 \leq 1/3
\]
and $\P(E^c) \leq 3n^2 e^{-cd}$,
where $c>0$ is an absolute constant. 
\end{lemma}
\begin{proof}
This follows from repeated application of Lemma \ref{lem:conc1} with $\epsilon = 1/100$ and a union bound.
\end{proof}

We can now show that gaussian vectors have the well-distributed property with high probability.  Recall that $S(x,y) = \Span(x,y)$.  

\begin{lemma} \label{lem:GT}
Let $t_1,\ldots t_n \in \mathbb{R}^d$ be i.i.d. $\mathcal{N}(0,\Id)$, and let $n\geq 16$ and $d\geq 3$.  For a fixed $k \neq l$, the inequality
\[
\sum_{i \in [n], i \neq l,k} \| P_{S(t_l-t_i, t_k-t_i)^\perp}(h) \|_2 \geq \frac{1}{5}(n-2)\|P_{S(t_l-t_k)^\perp}(h)\|_2
\]
holds for all $h \in \mathbb{R}^d$ with probability of failure at most $5n e^{-cd}$, where $c>0$ is an absolute constant.
\end{lemma}



\begin{proof}
Throughout the proof, constants named $c$ may be different from line to line, but are always bounded below by a positive absolute constant. For a fixed $(l,k)$, let $x = t_l, y = t_k$. We would like to show
\[
\sum_{i=1}^n \| P_{S(x-t_i, y-t_i)^\perp}(h) \|_2 \geq \frac{1}{5}n\|P_{S(x-y)^\perp}(h)\|_2
\]
We note that $S(x-t_i,y-t_i) = S(x-y, x+y -2t_i)$. Thus,
\[
P_{S(x-t_i, y-t_i)^\perp}(h) = P_{S(x-y, x+y-2t_i)^\perp}(h) = P_{S(x-y, x+y-2t_i)^\perp}(P_{S(x-y)^\perp}(h))
\]
Thus, it's enough to show
\[
\sum_{i=1}^n \| P_{S(x-y, x+y-2t_i)^\perp}(h) \|_2 \geq \frac{1}{5} n\|h\|_2
\]
for $h \perp (x-y)$. 

Now, for any vectors $v,w$, we have
\[
S(v,w) = S(v, w_{v^\perp})
\]
where $w_{v^\perp} = w - \langle w,\hat{v}\rangle \hat{v}$. If $h \perp v$, we have
\begin{align*}
P_{ S(v,w)^\perp}(h) &= P_{S(v, w_{v^\perp})^\perp}(h)\\
& = h - \langle h,\hat{v}\rangle \hat{v} - \langle h, \hat{w}_{v^\perp} \rangle \hat{w}_{v^\perp}\\
&= h - \left\langle h, \frac{w}{\|w_{v^\perp}\|_2} \right\rangle \frac{w_{v^\perp}}{\|w_{v^\perp}\|_2}\\
& = h - \langle h, \hat{w} \rangle \hat{w} + \langle h, \hat{w} \rangle \left[ \hat{w} - \frac{\|w\|_2}{\| w_{v^\perp} \|_2} \frac{w_{v^\perp}}{\|w_{v^\perp}\|_2} \right]\\
& = P_{S(w)^\perp}(h) + \langle h, \hat{w} \rangle z
\end{align*}
Where $z = \hat{w} - \frac{\|w\|_2}{\| w_{v^\perp} \|_2} \frac{w_{v^\perp}}{\|w_{v^\perp}\|_2}$. Now, assuming that $|\langle \hat{v} ,\hat{w} \rangle | < 1/2$ and using that
\[
\|w_{v^\perp}\|_2^2 = \|w - \langle w, \hat{v} \rangle \hat{v} \|_2^2 = \|w\|_2^2 - 2\|w\|_2^2 \langle \hat{w}, \hat{v} \rangle + \|w\|_2^2 |\langle \hat{w},\hat{v} \rangle|^2 \geq \|w\|_2^2(1 - 2 |\langle \hat{w}, \hat{v}\rangle|)
\]
we have
\begin{align*}
\|z\|_2 &= \left\| \hat{w} - \frac{\|w\|_2}{\|w_{v^\perp}\|_2^2} \left( w - \langle w, \hat{v} \rangle \hat{v}  \right) \right\|_2\\
&= \left\| \hat{w}\left[ 1 - \frac{\|w\|_2^2}{\|w_{v^\perp}\|_2^2}\right] + \frac{\|w\|_2^2}{\|w_{v^\perp}\|_2^2}\langle \hat{w}, \hat{v} \rangle \hat{v}  \right\|_2\\
& \leq \left | 1 - \frac{\|w\|_2^2}{\|w_{v^\perp}\|_2^2} \right | + \frac{\|w\|_2^2}{\|w_{v^\perp}\|_2^2} |\langle \hat{w}, \hat{v} \rangle | = \epsilon( \langle \hat{w},\hat{v}\rangle )\\
& = \frac{\|w\|_2^2}{\|w_{v^\perp}\|_2^2}\left(1 + |\langle \hat{w}, \hat{v} \rangle | \right) - 1\\
& \leq \frac{3 |\langle \hat{w}, \hat{v} \rangle |}{1- 2|\langle \hat{w}, \hat{v} \rangle |} \triangleq \zeta( \langle\hat{w},\hat{v} \rangle )
\end{align*}
Thus, we have
\[
\| P_{S(v,w)^\perp}(h) \|_2 \geq \|P_{S(w)^\perp}(h)\|_2 -  \zeta( \langle \hat{w},\hat{v} \rangle) \|h\|_2
\]
Therefore, by taking $v = x-y$ and $w = x+y -2t_i$, to conclude the desired statement of the present Lemma, it suffices to show that
\[
\sum_{i=1}^n \| P_{S(x+y-2t_i)^\perp}(h)\|_2 \geq \gamma n\|h\|_2
\]
where $\gamma > 1/5 + \zeta \left(\frac{\langle x-y, x+y + 2t_i \rangle}{\|x-y\|_2 \| x+y -2t_i \|_2}\right)$. Note that $x-y$ and $x+y - 2t_i$ are independent, and $\frac{1}{2}(x-y) =^d \frac{1}{6}(x+y-2t_i) =^d \mathcal{N}(0,I_{d\times d})$. Applying Lemma \ref{lem:conc1} to $x-y$ and $x+y-2t_i$ with a small enough value of $\epsilon$ to ensure $\zeta\left(\frac{\langle x-y, x+y + 2t_i \rangle}{\|x-y\|_2 \| x+y -2t_i \|_2}\right) < 1/20$, we get

\[
\P\left( \zeta\left(\frac{\langle x-y, x+y + 2t_i \rangle}{\|x-y\|_2 \| x+y -2t_i \|_2}\right) > \frac{1}{20} \right) \leq 3e^{-c d}
\]
Thus, it suffices to show with high probability, that
\[
\sum_{i=1}^n \| P_{S(x+y-2t_i)^\perp}(h)\|_2 \geq 0.3 n\|h\|_2,
\]
which we proceed to establish below. 

To begin, redefine $v,w$ as $v = x+y$ and $w = -2t_i$ and consider

\begin{align*}
\sum_{i=1}^n \left \| P_{S(v+w_i)^\perp}(h)\right \|_2 &\geq \left \|  \sum_{i=1}^n P_{S(v+w_i)^\perp}(h)\right\|_2\\
&= \left \| \sum_{i=1}^n \left(h - \frac{1}{\|v+w_i\|_2^2}\langle h, v+ w_i \rangle (v+ w_i) \right)\right \|_2\\
&\geq n\|h\|_2 - \left \| \sum_{i=1}^n \frac{1}{\|v+w_i\|_2^2} (v+ w_i)(v+ w_i)^* h\right \|_2  \\
&\geq n\|h\|_2 - \left \| \sum_{i=1}^n \frac{1}{\|v+w_i\|_2^2} (v+ w_i)(v+ w_i)^* \right\|_{\text{op}} \|h\|_2\\
& \geq \|h\|_2 \left[n - \frac{1}{\min_{i}\|v+w_i\|_2^2} \left \| \sum_{i=1}^n (v+ w_i)(v+ w_i)^* \right\|_{\text{op}} \right]
\end{align*}
where in the last inequality we used 
\[
\sum_{i=1}^n \frac{1}{\|v+w_i\|_2^2} (v+ w_i)(v+ w_i)^* \preceq \frac{1}{\min_{i}\|v+w_i\|_2^2}  \sum_{i=1}^n (v+ w_i)(v+ w_i)^*
\]
Now, let $A = \sum_{i=1}^n e_i w_i^* \in \mathbb{R}^{n\times d}$. We have  
\begin{align*}
\left \| \sum_{i=1}^n (v+ w_i)(v+ w_i)^* \right\|_{\text{op}} &= \left \| \sum_{i=1}^n (vv^* + vw_i^* + w_i v^* + w_i w_i^*) \right\|_{\text{op}}\\
& \leq n\|vv^*\|_{\text{op}} + \left\| v \left( \sum_{i=1}^n w_i \right)^* +  \left( \sum_{i=1}^n w_i \right)v^* \right\|_{\text{op}} + \left \| \sum_{i=1}^n w_i w_i^* \right\|_{\text{op}}\\
& \leq n \|v\|_2^2 + 2\|v\|_2\left \| \sum_{i=1}^n w_i \right\|_2 + \left \| \sum_{i=1}^n w_i w_i^* \right\|_{\text{op}}\\
& = n \|v\|_2^2 + 2\|v\|_2\left \| \sum_{i=1}^n w_i \right\|_2 + \sigma_{\max}(A)^2\\
\end{align*}

Thus,
\[
\sum_{i=1}^n \left \| P_{S(v+w_i)^\perp}(h)\right \|_2 \geq \|h\|_2\left[n - \frac{n \|v\|_2^2 + 2\|v\|_2\left \| \sum_{i=1}^n w_i \right\|_2 + \sigma_{\max}(A)^2}{\min_{i}\|v+w_i\|_2^2} \right]
\]

Now, consider the event
\[
E = \left\{ \min_{i} \|v+w_i\|_2^2 \geq 6d\beta_1, \quad \|v\|_2^2 \leq 2d\beta_2, \quad \left \| \sum_{i=1}^n w_i \right\|_2^2 \leq 4nd\beta_3, \quad \sigma_{\max}(A)^2 \leq n\beta_4  \right\}
\]
On E, we have
\begin{align*}
\sum_{i=1}^n \left \| P_{S(v+w_i)^\perp}(h)\right \|_2 &\geq \|h\|_2 \left[ n - \frac{1}{6d\beta_1}\left( 2nd\beta_2 + 2\sqrt{2d\beta_2}2\sqrt{nd}\sqrt{\beta_3} + n\beta_4  \right) \right]\\
&= \|h\|_2 \left[n - \frac{1}{3}n \frac{\beta_2}{\beta_1} - \frac{4\sqrt{2}d\sqrt{n} \sqrt{\beta_2\beta_3}}{6d\beta_1}-\frac{\beta_4 }{6d\beta_1}n \right]\\
& = \|h\|_2 \left[n\left(1 - \frac{1}{3} \frac{\beta_2}{\beta_1} - \frac{\beta_4}{6d\beta_1}- \frac{1}{\sqrt{n}}\frac{ 4\sqrt{2} \sqrt{\beta_2\beta_3}}{6\beta_1} \right)\right]\\
\end{align*}

Now, note that $ \frac{1}{6}\|v+w_i\|_2^2 =^d \frac{1}{2}\|v\|_2^2 =^d \frac{1}{4n} \left \| \sum_{i=1}^n w_i \right\|_2^2 =^d \chi^2(d)$ and $A$ is a random $n\times d$ matrix with i.i.d. $\mathcal{N}(0,1)$ entries. 

Thus by applying Lemma \ref{lem:conc1} we have 
\[
\P\left( 6d(1-\epsilon) \leq \|v+w_i\|_2^2 \leq 6d(1+\epsilon) \right) \geq 1 - e^{-c \epsilon^2 d}
\]

\[
\P\left( 2d(1-\epsilon) \leq \|v\|_2^2 \leq 2d(1+\epsilon) \right) \geq 1 - e^{-c \epsilon^2 d}
\]

\[
\P\left( 4nd(1-\epsilon) \leq \left \| \sum_{i=1}^n w_i \right\|_2^2 \leq 4nd(1+\epsilon) \right) \geq 1 - e^{-c \epsilon^2 d}
\]

where $c>0$ is a universal constant. Also by taking $t = \sqrt{2d}$ in Lemma \ref{lem:conc2} we get
\[
\P \left( \sigma_{\text{max}}(A) \geq \sqrt{n} +2\sqrt{d} \right) \leq 2e^{-d}
\]

Now, let $\beta_1 = 1-\frac{1}{100}, \quad \beta_2 = \beta_3 = 1 + \frac{1}{100}, \quad \beta_4 = \frac{d}{2}$, which gives
\[
\frac{1}{3} \frac{\beta_2}{\beta_1} \leq 1/3 +1/99, \quad \frac{1}{\sqrt{n}}\frac{ 4\sqrt{2} \sqrt{\beta_2\beta_3}}{6\beta_1} < \frac{1}{\sqrt{n}}, \quad \frac{\beta_4}{5d} = 1/10
\]

We have
\[
\P  \left(\sigma_{\max} (A) \geq \sqrt{n \beta_4}\right) \leq \P\left( \sigma_{\max} (A) \geq \sqrt{n} + 2\sqrt{d} \right) \leq 2e^{-d}
\]
whenever $\sqrt{n} + 2\sqrt{d} \leq \sqrt{n}\sqrt{d/2}$, which holds whenever
\[
n \geq \left( \frac{2\sqrt{d}}{\sqrt{d/2} -1} \right)^2
\]
which holds for $n \geq 16$ when $d\geq 3$. Since for $n\geq 16$, $\frac{1}{\sqrt{n}} \leq 1/4$, we have on E
\[
\sum_{i=1}^n \left \| P_{S(v+w_i)^\perp}(h)\right \|_2 \geq 0.3 n \|h\|_2
\]
Thus,
\[
\P\left ( \sum_{i=1}^n \left \| P_{S(v+w_i)^\perp}(h)\right \|_2 < 0.3 n \|h\|_2 \right) \leq \P(E^c) \leq (n+3)e^{-cd}
\]
where $c>0$ is an absolute constant. 

Combining all of the above, we get
\[
\sum_{i=1}^n \| P_{S(x-t_i, y-t_i)^\perp}(h) \|_2 \geq \frac{1}{5}n\|P_{S(x-y)^\perp}(h)\|_2
\]

with probability of failure at most $5ne^{-cd}$. 


 \end{proof}

\begin{lemma} \label{lem:ptyp-well}
Let $G([n],E)$ be $p$-typical, and $t_1,\ldots t_n \sim \mathcal{N}(0,I_{d\times d})$ be i.i.d. Then $T = \{t_i\}_{i \in [n]}$ is $\frac{1}{5}$-well distributed along $G$ with probability at least $1-10n^3 e^{-cd}$, where $c>0$ is an absolute constant. 
\end{lemma}
\begin{proof}
For each $ij \in E$, let $S_{ij} = \{ k \in [n]; ik, jk \in E(G) \}$ and note that $|S_{ij}| \leq 2np^2$. Now apply Lemma \ref{lem:GT} to the set of vectors $ \{t_i, t_j\} \bigcup \{ t_k \}_{k \in \mathcal{I}_{ij}}$, with the distinguished vectors being $\{t_i, t_j\}$, which gives the desired property for the pair $(i,j)$ with probability of failure at most $5(|S_{ij}|)e^{-cd} \leq 5 (2np^2) e^{-cd}$, where $c>0$ is an absolute constant. Taking the union bound over pairs of distinct integers $i,j \in [n]$, we get the desired property simultaneously for all pairs with probability at least $1- n^2 \cdot 5(2np^2) e^{-cd} = 1 - 10n^3 p^2 e^{-cd} \geq 1- 10n^3 e^{-cd}$.
\end{proof}

\subsection{Random graphs are $p$-typical with high probability} \label{sec-random-graph}

\begin{lemma} \label{lem:ptyp}
There exists an absolute constant $c > 0$ such that for all positive real numbers $p \le 1$, $G(n,p)$ is $p$-typical with probability at least $1-n^2e^{-cnp^2}$ if $np \ge 4 \log n$.
\end{lemma}
\begin{proof}
A graph is not connected only if there exists a partition $V_1 \cup V_2$ of its vertex
set for which there are no edges between $V_1$ and $V_2$. Without loss of generality,
we may assume that $|V_1| \le \lfloor \frac{n}{2} \rfloor$. Since the number
of ways to choose a set of size $k$ from a set of size $n$ is ${n \choose k}$, 
the probability that $G(n,p)$ is not connected is at most 
\[
	\sum_{k=1}^{\lfloor n/2 \rfloor} {n \choose k} (1-p)^{k(n-k)}
	\le \sum_{k=1}^{\lfloor n/2 \rfloor} \left(\frac{en}{k}\right)^{k} e^{-pk(n-k)}
	< \sum_{k=1}^{\lfloor n/2 \rfloor} \left(n e^{1-p(n-k)} \right)^{k}.
\]
Since $k \le \lfloor \frac{n}{2} \rfloor$, we have
$n e^{1-p(n-k)} \le n e^{1-pn/2} < 1$ (since $np \ge 4 \log n$). 
Therefore the summand on the right-hand-side is at most $(ne^{1-pn/2})^{k}$,
which is maximized at $k=1$. 
This shows that the probability that $G(n,p)$ is not connected is at most $n^2 e^{1-pn/2}$.

In $G(n,p)$, for a fixed vertex $v$, the expected value of $\deg(v)$is
$(n-1)p$, and for a pair of vertices $v,w$, the expected value of
the codegree of $v$ and $w$ is $(n-2)p^{2}$. Therefore the lemma
follows from Chernoff's inequality --- see Fact 4 from  \cite{AngluinValiant} --- and a union bound. 
\end{proof}



\subsection{Proof of Theorem \ref{thm-complete}} \label{sec-proof-random-theorem}

We can now prove the high dimensional recovery theorem, which we state here again for convenience:

\addtocounter{theorem}{-3}

\begin{theorem} 
Let $G([n],E)$ be drawn from $G(n,p)$ for some $p = \Omega(n^{-1/4})$. Take $ \tnot_1, \ldots \tnot_n \sim \mathcal{N}(0, I_{d \times d})$ to be i.i.d., independent from $G$. There exists an absolute constant $c>0$ and a $\gamma = \Omega(p^4)$ not depending on $n$, such that if $\max(\frac{2^6}{c^6}, \frac{4^3}{c^3} \log^3 n) \leq  n \leq e^{\frac{1}{6}c d}$ and $d = \Omega(1)$, then there exists an event with probability at least  $1- e^{-n^{1/6}} - 13 e^{-\frac{1}{2}c d}$, on which the following holds:\\[1em]
For arbitrary  subgraphs $\Eb$ satisfying $\max_i \deg_b(i) \leq \gamma n$ and arbitrary pairwise direction corruptions  $\vij \in \mathbb{S}^{d-1}$ for $ij \in \Eb$,  the convex program \eqref{shapefit} has a unique minimizer equal to $\left \{\alpha \Bigl(\toi - \tnotbar \Bigr)\right\}_{i \in [n]}$ for some positive $\alpha$ and for $\tnotbar = \frac{1}{n}\sum_{i\in [n]} \toi$. 
\end{theorem}
\begin{proof}
It is enough to verify that $G$, $T$ and $E_b$ in the assumption of the present theorem satisfy the deterministic conditions 1--6 in Theorem 2, with appropriate constants $p, \beta, c_0, \epsilon, c_1$, and with the purported probability. By Lemma \ref{lem:ptyp}, Lemma \ref{lem:conc3}, and Lemma \ref{lem:ptyp-well}, we have that Condition 1 holds with value $p$, Condition 2 holds with $\beta = \sqrt{\frac{2}{3}}$, Condition 3 holds with $c_0 = \frac{9}{10}$, and Condition 5 holds with $c_1 = \frac{1}{5}$, with probability at least
\[
1-n^2 e^{-cnp^2} - 3n^2 e^{-cd} - 10n^3 e^{-cd}
\]
where $c>0$ is an absolute constant.

Thus, taking any $E_b$, which satisfies Condition 4 with $\gamma = \frac{p^4}{10^7} \leq \frac{\beta c_0 c_1^2 p^4}{256\cdot 32\cdot 64 \cdot 3}$ , we get that recovery via ShapeFit is guaranteed. Note that the condition $\max \deg_b(i) \leq \gamma n$ is nontrivial when $p = \Omega(n^{-1/4})$.  Using the requirements on $n$ and $p$, we have $n^2 e^{-cnp^2} \leq n^2 e^{-cn^{1/3}} \leq e^{-\frac{1}{6}n}$ and $13n^3 e^{-cd} \leq 13(e^{\frac{1}{6}cd})^3 e^{-cd} \leq 13 e^{-\frac{1}{2}cd}$. Thus, the probability of exact recovery via ShapeFit, uniformly in $E_b$ and $v_{ij}$ satisfying the assumptions of the theorem, is at least
\[
1- e^{-n^{1/6}} - 13 e^{-\frac{1}{2}c d}. \qedhere
\]


\end{proof}

\section{Proof of three-dimensional recovery} \label{sec-proofs-three-d}

The proof of recovery in three dimensions parallels the proof in high dimensions, but it is more technical because it can not capitalize on the concentration of measure phenomenon in high dimensions.
Specifically, the additional technicality in three dimensions comes from the fact that
for large $n$, there exist pairs of locations $\toi$, $\toj$
that are close to each other, i.e., $\|\toij\|_2$ is small.
For such pairs of vectors, with high probability, for all $k \neq i,j$
the value of $1 - \langle \toikhat, \tojkhat \rangle^2$ will be small.
This fact introduces the following two main obstacles in carrying out the same analysis:
\begin{enumerate}
  \setlength{\itemsep}{1pt} \setlength{\parskip}{0pt}
  \setlength{\parsep}{0pt}
  \item There is no uniform lower bound on  $1 - \langle \toikhat, \tojkhat \rangle^2$.  Hence Condition 2 in Theorem~\ref{thm:mainthm} fails.
  \item There is no uniform lower bound on $\|t_{ij}^{(0)}\|_2$. Hence Condition 3 in Theorem~\ref{thm:mainthm} fails. 
\end{enumerate}
These are indeed obstacles since
the gains in rotational motions coming from Lemmas~\ref{lem:rigidity1} and \ref{lem:rigidity2}
are proportional to $\sqrt{1 - \langle \toikhat, \tojkhat \rangle^2}$ and $\| \toij \|_2$.
We avoid these difficulties and prove the three-dimensional analogue of Theorem~\ref{thm:mainthm} by weakening Conditions 2 and 3. 
Roughly speaking, in $\mathbb{R}^3$, Condition 2 holds for most triples $i,j,k \in [n]$ (instead of all triples)
and Condition 3 gets replaced by a one-sided version where we only have a uniform  upper bound on the lengths $\| \toij \|_2$.


Unlike in the high-dimensional case where we allowed a constant fraction of edges incident to each
vertex to be corrupted, the three-dimensional case requires the fraction of corrupted edges incident 
to each vertex to be at most $O(\frac{1}{\log^3 n})$.
This additional poly-logarithmic factor is due to the fact that our well-distributedness proof in three dimensions hinges on the maximum $\ell_2$ norm of locations, which is $\Omega(\sqrt{\log n})$ with high probability.
It can be removed for a distribution of locations that has a uniform constant upper bound on $\|\toi\|_2$.

\subsection{Deterministic recovery theorem in three dimensions} \label{sec-deterministic-theorem-three-d}


We now state deterministic conditions on the graph $G$, the corrupted observations $\Eb$, and the locations $\Tnot$ that guarantee recovery.
Recall the definition $\mu=\frac{1}{|E(G)|}\sum_{ij\in E(G)}\|t_{ij}^{(0)}\|_2$. 

\setcounter{theorem}{3}
\begin{theorem} \label{thm-deterministic-three-d}
Suppose $\Tnot, \Eb, G$ satisfy the conditions
\begin{enumerate}
  \setlength{\itemsep}{1pt} \setlength{\parskip}{0pt}
  \setlength{\parsep}{0pt}
	\item \label{enu:typical}The underlying graph $G$ is $p$-typical,
	\item \label{enu:angle}For all distinct $i,j\in[n]$, for all but at most $\eps_{1}n$ indices $k\in[n]$ satisfying $k \neq i,j$, we have $1-\langle\hat{t}_{ij},\hat{t}_{ik}\rangle^{2}\ge\beta^{2}$ and $1-\langle\hat{t}_{ij},\hat{t}_{jk}\rangle^{2}\ge\beta^{2}$,
	\item \label{enu:length}For all distinct $i,j\in[n]$, we have $\|t_{ij}^{(0)}\|_2\le c_{0}\mu$, 	\item \label{enu:bad}Each vertex has at most $\eps_0 n$ edges in $\Eb$ incident to it,
	\item \label{enu:welldistributed}The set $\{t_{i}^{(0)}\}_{i \in [n]}$ is $c_{1}$-well-distributed
along $G$,
	\item \label{enu:general}No three vectors $\toi, \toj, \tok$ are collinear for distinct $i,j,k$.
\end{enumerate}
for constants $0< p, \beta, \eps_0, \eps_1, c_{1} \leq 1 \le c_0$.  If $\eps_0 \leq \frac{\beta c_1^2 p^4}{32 \cdot 3 \cdot 64 \cdot 1024 c_0^2}$ and $\eps_1 \le \frac{p}{192c_0}$, then $L(\Tnot)\neq 0$ and $\Tnot / L(\Tnot)$ is the unique optimizer of ShapeFit.
\end{theorem}

Note that all six 
conditions are invariant under translation and non-zero scalings of $\Tnot$ (Condition~\ref{enu:length}
is invariant since both $t_{ij}^{(0)}$ and $\mu$ scale together and are invariant under translation).
Before we prove the theorem, we establish that $L(\Tnot)\neq 0$ when $\eps_0$ is small.  This non-equality guarantees that some scaling of $\Tnot$ is feasible whenever, roughly speaking, $|\Eb| < |\Eg|$.

\begin{lemma} \label{lem:T0-feasible-three-d} 
If $\eps_0 < \frac{p}{8c_0}$, then $L(\Tnot) \neq 0$.
\end{lemma}
\begin{proof}
Since $v_{ij} = \toijhat$ for all $ij \in E_g$, we have
\[
	L(\Tnot) = \sum_{ij \in E(G)} \langle \toij, v_{ij} \rangle 
	\geq \sum_{ij \in E_g} \| \toij\|_2 - \sum_{ij \in E_b} \| \toij \|_2
	= \sum_{ij \in E(G)} \| \toij\|_2 - 2\sum_{ij \in E_b} \| \toij \|_2.
\]
By Condition \ref{enu:length},  $\sum_{ij \in E_b} \| \toij \|_2 \le c_0 \mu |E_b| \le c_0\mu \cdot \eps_0n^2 < \frac{1}{8}n^2p \mu$. Since Condition \ref{enu:typical} implies $|E(G)| \ge \frac{1}{4}n^2p$, we have
$\sum_{ij \in E(G)} \| \toij\|_2 \ge \frac{1}{4}n^2p\mu$. Therefore it follows that $L(\Tnot) > 0$. 
\end{proof}


\subsection{Proof of Theorem \ref{thm-deterministic-three-d}} \label{sec-proof-deterministic-3d}

Lemmas \ref{lem:rigidity1} and \ref{lem:rigidity2} will be repeatedly
used throughout the proof. Note that these lemmas can be used only
if the given set of vectors satisfies a certain condition on the angles between
them. For each distinct $ij\in E(K_{n})$, define $B(ij)$ as the
set of edges $k\ell\in E(K_{n})$ such that $\sqrt{1-\langle\hat{t}_{ac}^{(0)},\hat{t}_{bc}^{(0)}\rangle^{2}}<\beta$
holds for some distinct $a,b,c\in\{i,j,k,\ell\}$ satisfying
$(a,b)\neq(i,j)$. Note that Lemmas~\ref{lem:rigidity1} and \ref{lem:rigidity2} can be applied to the
set of indices $\{i,j,k,\ell\}$ (having size either 3 or 4) for all $k\ell \notin B(ij)$.
The following lemma shows that $B(ij)$ is small for each $ij$.

\begin{lemma}
\label{lem:badK4}For each $ij\in E(K_{n})$, we have $|B(ij)|\le6\varepsilon_{1}n^{2}$.
\end{lemma}
\begin{proof}
For each $ab\in E(K_{n})$, define $B_{3}(ab)$ as the set of indices
$c\in[n]$ distinct from $a,b$ for which $\sqrt{1-\langle \hat{t}_{ab}^{(0)}, \hat{t}_{ac}^{(0)}\rangle^{2}}<\beta$
or $\sqrt{1-\langle\hat{t}_{ab}^{(0)},\hat{t}_{bc}^{(0)}\rangle^{2}}<\beta$
holds. Condition \ref{enu:angle} implies $|B_{3}(ab)|\le\varepsilon_{1}n$
for all $ab \in E(K_n)$. One can check that $k\ell\in B(ij)$ if and only if
one of the following events hold: $k\in B_{3}(ij)$, $\ell\in B_{3}(ij),$
$k\in B_{3}(i\ell)\cup B_{3}(j\ell)$, $\ell\in B_{3}(ik)\cup B_{3}(jk)$.
Therefore 
\begin{eqnarray*}
|B(ij)| & \le & 2|B_{3}(ij)|\cdot n+\sum_{\ell\neq i,j}\Big(|B_{3}(i\ell)|+|B_{3}(j\ell)|\Big)+\sum_{k\neq i,j}\Big(|B_{3}(ik)|+|B_{3}(jk)|\Big)\\
 & \le & 2\varepsilon_{1}n^{2}+n\cdot2\varepsilon_{1}n+n\cdot2\varepsilon_{1}n=6\varepsilon_{1}n^{2}.\qedhere
\end{eqnarray*}
\end{proof}

We now prove the deterministic recovery theorem in three dimensions.

\begin{proof}[Proof of Theorem \ref{thm-deterministic-three-d}]

By Lemma \ref{lem:T0-feasible-three-d} and the fact that Conditions 1--6 are invariant under global translation and nonzero scaling, we can take $\Tnotbar=0$ and $L(\Tnot) = 1$ without loss of generality.  The variable $\mu$ is to be understood accordingly.  

We will directly prove that $R(T) > R(\Tnot)$ for all $T\neq \Tnot$ such that $L(T)=1$ and $\Tbar = 0$. Consider an arbitrary feasible $T$ and recall the notation $\tij = (1 + \deltaij) \toij + \etaij \sij$ where $\sij$ is a unit vector orthogonal to $\toij$ and $\etaij = \| P_{\toijperp} \tij \|_2$. A useful lower bound for the objective $R(T)$ is given by 
\begin{eqnarray}
	R(T)
	=\sum_{ij \in E(G)}\|\proj_{v_{ij}^{\perp}}t_{ij}\| _2
	& = & \sum_{ij\in E_{g}}\eta_{ij}+\sum_{ij\in E_{b}}\|\proj_{v_{ij}^{\perp}}t_{ij}\|_2\nonumber \\
	& \ge & \sum_{ij\in E_{g}}\eta_{ij}+\sum_{ij\in E_{b}}\left(\|\proj_{v_{ij}^{\perp}}t_{ij}^{(0)}\|_2-|\delta_{ij}|\|t_{ij}^{(0)}\|_2-\eta_{ij}\right)\nonumber \\
	& = & R(\Tnot)+\sum_{ij\in E_{g}}\eta_{ij}-\sum_{ij\in E_{b}}(|\delta_{ij}|\|t_{ij}^{(0)}\|_2+\eta_{ij}).\label{eq:gain3-3d}
\end{eqnarray}

Suppose that $\sum_{ij\in E_{b}}|\delta_{ij}|\|t_{ij}^{(0)}\|_2<\sum_{ij\in E_{b}}\eta_{ij}$.
Since $\varepsilon_0 \le\frac{c_{1}p^{2}}{16}$, Lemma \ref{lem:badgood}
implies $\sum_{ij\in E_{b}}\eta_{ij}\le\frac{1}{2}\sum_{ij\in E_{g}}\eta_{ij}$.
Therefore by (\ref{eq:gain3-3d}), we have
\begin{eqnarray*}
R(T) & \ge & R(T_{0})+\sum_{ij\in E_{g}}\eta_{ij}-\sum_{ij\in E_{b}}(|\delta_{ij}|\|t_{ij}^{(0)}\|_2+\eta_{ij})\\
 & > & R(T_{0})+\sum_{ij\in E_{g}}\eta_{ij}-\sum_{ij\in E_{b}}2\eta_{ij}\ge R(T_{0}).
\end{eqnarray*}
Hence we may assume 
\begin{equation}
\sum_{ij\in E_{b}}|\delta_{ij}|\|t_{ij}^{(0)}\|_2\ge\sum_{ij\in E_{b}}\eta_{ij}.\label{eq:badtwist-3d}
\end{equation}
In other words, the total parallel motion is larger than the total rotational
motions on the bad edges. The key idea of the proof is to show that
parallel motions on bad edges induce a large amount of rotational motions
on good edges. 

In the case $|\Eb| \neq 0$,  define $\overline{\delta}:=\frac{\sum_{ij\in E_{b}}|\delta_{ij}|\|t_{ij}^{(0)}\|_2}{\sum_{ij\in E_{b}}\|t_{ij}^{(0)}\|_2}$
as the average `relative parallel motion' on the bad edges. 
For distinct $ij,k\ell\in E(K_{n})$, if $\{i,j\}\cap\{k,\ell\}=\emptyset$,
then define $\eta(ij,k\ell)=\eta_{ij}+\eta_{ik}+\eta_{i\ell}+\eta_{jk}+\eta_{j\ell}+\eta_{k\ell}$,
and if $\{i,j\}\cap\{k,\ell\}\neq\emptyset$ (without loss of generality,
assume $\ell=i$), then define $\eta(ij,k\ell)=\eta_{ij}+\eta_{ik}+\eta_{jk}$. 

\medskip{}

\noindent \textbf{Case 0}. $\bar{\delta} = 0$ or $|\Eb| = 0$.
\medskip

Note that $\bar{\delta}=0$ implies $\delta_{ij} =0$ for all $ij \in E_b$, which by \eqref{eq:badtwist} implies
$\eta_{ij} =0$ for all $ij \in E_b$.  Therefore by \eqref{eq:gain3}, we have
\[
	R(T) \geq R(\Tnot) + \sum_{ij \in E_g} \eta_{ij}.
\]
If $\sum_{ij \in E_g} \eta_{ij} > 0$, then we have $R(T) > R(\Tnot)$. Thus 
we may assume that $\eta_{ij} = 0$ for all $ij \in E_g$. In this case, we will show that $T = \Tnot$.

By Lemma~\ref{lem:transference}, if $\varepsilon_0 \le \frac{c_1p^2}{8}$, then
$\eta_{ij}=0$ for all $ij \in E(G)$ implies that
$\eta_{ij} = 0$ for all $ij \in E(K_n)$. 
For $ij \in E_b$, since $\delta_{ij} = \eta_{ij} = 0$, it follows that $\ell_{ij} = \ell_{ij}^{(0)}$.
Since $\delta_{ij} = \ell_{ij} - \ell_{ij}^{(0)}$ for $ij \in E_g$, we have
\[
	0 = \sum_{ij \in E(G)} (\ell_{ij} - \ell_{ij}^{(0)})
	= \sum_{ij \in E_b} (\ell_{ij} - \ell_{ij}^{(0)}) + \sum_{ij \in E_g} (\ell_{ij} - \ell_{ij}^{(0)})
	= \sum_{ij \in E_g} (\ell_{ij} - \ell_{ij}^{(0)})
	= \sum_{ij \in E_g} \delta_{ij}\|\toij\|_2,
\] 
where the first equality is because $L(T) = L(\Tnot) = 1$.
By Condition 0, we have $\|t_{ij}^{(0)}\| \neq 0$ for all $ij \in E_g$.
Hence if $\delta_{ij} \neq 0$ for some $ij \in E_g$, then there exists $ab, cd \in E_g$ 
such that $\delta_{ab} > 0$ and $\delta_{cd} < 0$. 
By Lemma~\ref{lem:rigidity1} or \ref{lem:rigidity2} and Condition 6, 
this forces $\eta(ab,cd) > 0$, contradicting the fact that
$\eta_{ij} = 0$ for all $ij \in E(K_n)$. Therefore $\delta_{ij} = 0$ for all $ij \in E_g$, and hence
$\delta_{ij} = 0$ for all $ij \in E(G)$. 

Define $t_i = t_i^{(0)} + h_i$ for each $i \in [n]$. 
Because $\eta_{ij} = \delta_{ij} = 0$ for all $ij \in E(G)$, we have $h_i = h_j$ 
for all $ij \in E(G)$. Since $G$ is connected (by Condition~\ref{enu:typical}), this implies $h_i = h_j$ for all 
$i \in [n]$. Then by the constraint $\sum_{i \in [n]} t_i = \sum_{i \in [n]} t_i^{(0)} = 0$, 
we get $h_i = 0$ for all $i \in [n]$. Therefore $T = \Tnot$.
This proves Case 0. \\

\medskip

We may now assume that $\overline{\delta}\neq0$.
Since $\ell_{ij}-\ell_{ij}^{(0)}=\delta_{ij}\|t_{ij}^{(0)}\|_2$ for
$ij\in E_{g}$, we have 
\begin{eqnarray*}
0=\sum_{ij \in E(G)}(\ell_{ij}-\ell_{ij}^{(0)}) & = & \sum_{ij\in E_{b}}(\ell_{ij}-\ell_{ij}^{(0)})+\sum_{ij\in E_{g}}\delta_{ij}\|t_{ij}^{(0)}\|_2.
\end{eqnarray*}
Therefore 
\begin{eqnarray}
\left|\sum_{ij\in E_{g}}\delta_{ij}\|t_{ij}^{(0)}\|_2\right| & \le & \left|\sum_{ij\in E_{b}}(\ell_{ij}-\ell_{ij}^{(0)})\right|\le\sum_{ij\in E_{b}}(|\delta_{ij}|\|t_{ij}^{(0)}\|_2+\eta_{ij})\nonumber \\
 & \le & 2\sum_{ij\in E_{b}}|\delta_{ij}|\|t_{ij}^{(0)}\|_2.\label{eq:good_parallel}
\end{eqnarray}
where the final inequality follows from (\ref{eq:badtwist}). 

Define $E_{g}'=\{ij\in E_{g}\,:\,\|t_{ij}^{(0)}\|_2\ge\frac{1}{2}\mu\}$
as the set of `long' good edges. Since $\sum_{ij\in E_{g}\setminus E_{g}'}\|t_{ij}^{(0)}\|_2 < \frac{1}{2}\mu|E_{g}|$,
we have 
\begin{eqnarray*}
	\sum_{ij\in E_{g}'}\|t_{ij}^{(0)}\|_2 
	& = & \sum_{ij\in E(G)}\|t_{ij}^{(0)}\|_2-\sum_{ij\in E_{b}}\|t_{ij}^{(0)}\|_2-\sum_{ij\in E_{g}\setminus E_{g}'}\|t_{ij}^{(0)}\|_2\\
	& > & \mu|E(G)|-c_{0}\mu\cdot|E_{b}|-\frac{1}{2}\mu|E_{g}| \geq \mu\cdot\frac{1}{16}n^{2}p.
\end{eqnarray*}
where the last inequality uses $|E(G)| \geq \frac{n^2 p}{4}$, $|\Eb| < \eps_0 n^2$, $|\Eg| \leq |E(G)|$, $\eps_0 < \frac{p}{16 c_0}$.
By Condition~\ref{enu:length}, we have $\|t_{ij}^{(0)}\|_2\le c_{0}\mu$
for all $ij$, and thus it follows that 
\begin{equation}\label{eq:longgoodedges}
|E_{g}'|\ge\frac{1}{16c_{0}}n^{2}p.
\end{equation}

\medskip

\noindent\textbf{Case 1}. $\overline{\delta} \neq 0$ and  $\sum_{ij\in E_{g}'}|\delta_{ij}|<\frac{1}{8}\overline{\delta}|E_{g}'|$ and $|\Eb| \neq 0$.

In this case, we will exploit the fact that there is a difference between average
relative parallel motions on long good edges and that on bad edges, to show that there is
a large amount of rotational motion on the $K_{4}$s of the form $\{i,j,k,\ell\}$
where $ij\in E_{b}$ and $k\ell\in E_{g}'$. Define $L_{b}=\{ij\in E_{b}:|\delta_{ij}|\ge\frac{1}{2}\overline{\delta}\}.$
Note that $\sum_{ij\in E_{b}\setminus L_{b}}|\delta_{ij}|\|t_{ij}^{(0)}\|_2<\frac{1}{2}\overline{\delta}\sum_{ij\in E_{b}}\|t_{ij}^{(0)}\|_2=\frac{1}{2}\sum_{ij\in E_{b}}|\delta_{ij}|\|t_{ij}^{(0)}\|_2$.
Therefore 
\begin{eqnarray}
\sum_{ij\in L_{b}}|\delta_{ij}|\|t_{ij}^{(0)}\|_2 & = & \sum_{ij\in E_{b}}|\delta_{ij}|\|t_{ij}^{(0)}\|_2-\sum_{ij\in E_{b}\setminus L_{b}}|\delta_{ij}|\|t_{ij}^{(0)}\|_2>\frac{1}{2}\sum_{ij\in E_{b}}|\delta_{ij}|\|t_{ij}^{(0)}\|_2.\label{eq:large_delta_3d}
\end{eqnarray}
Define $F_{g}=\{ij\in E_{g}:|\delta_{ij}|<\frac{1}{4}\overline{\delta}\}$.
Then by the condition of Case 1,
\[
\frac{1}{8}\overline{\delta}|E_{g}'|>\sum_{ij\in E_{g}'}|\delta_{ij}|\ge\sum_{ij\in E_{g}'\setminus F_{g}}|\delta_{ij}|\ge\frac{1}{4}\overline{\delta}|E_{g}'\setminus F_{g}|,
\]
and therefore $|E_{g}'\setminus F_{g}|<\frac{1}{2}|E_{g}'|$, or equivalently,
$|F_{g}|>\frac{1}{2}|E_{g}'|\ge\frac{1}{32c_{0}}n^{2}p$ (where the
second inequality comes from (\ref{eq:longgoodedges})).

For each $ij\in L_{b}$ and $k\ell\in F_{g}\setminus B(ij)$, by Lemmas
\ref{lem:rigidity1} and \ref{lem:rigidity2}, we have $\eta(ij,k\ell)\ge\frac{\beta}{4}|\delta_{k\ell}-\delta_{ij}|\|t_{ij}^{(0)}\|_2\ge\frac{\beta}{4}\cdot\frac{1}{2}|\delta_{ij}|\|t_{ij}^{(0)}\|_2$.
Therefore, 
\begin{eqnarray*}
\sum_{ij\in E_{b}}\sum_{k\ell\in E_{g}}\eta(ij,k\ell) & \ge & \sum_{ij\in L_{b}}\sum_{k\ell\in F_{g}\setminus B(ij)}\frac{\beta}{8}|\delta_{ij}|\|t_{ij}^{(0)}\|_2=\sum_{ij\in L_{b}}|F_{g}\setminus B(ij)|\cdot\frac{\beta}{8}|\delta_{ij}|\|t_{ij}^{(0)}\|_2.
\end{eqnarray*}
By Lemma \ref{lem:badK4}, we know that $|B(ij)|<6\varepsilon_{1}n^{2}$
holds for all $ij\in E(K_{n})$. For $\varepsilon_1 \le \frac{p}{192c_0}$, we have
\[
	|F_{g}\setminus B(ij)| 
	> \frac{1}{32c_{0}}n^{2}p-6\varepsilon_{1}n^{2}
	\ge\frac{1}{64c_{0}}n^{2}p.
\]
Therefore
\[
	\sum_{ij\in E_{b}}\sum_{k\ell\in E_{g}}\eta(ij,k\ell)
	> \frac{\beta}{8}\cdot\frac{1}{64c_{0}}n^{2}p\cdot\sum_{ij\in L_{b}}|\delta_{ij}|\|t_{ij}^{(0)}\|_2\ge\frac{\beta}{1024c_{0}}n^{2}p\sum_{ij\in E_{b}}|\delta_{ij}|\|t_{ij}^{(0)}\|_2,
\]
where the second inequality comes from (\ref{eq:large_delta_3d}).

For each $ij\in E(K_{n})$, we would like to count how many times
each $\eta_{ij}$ appear on the left hand side. If $ij\in E_{b}$,
then there are at most ${n \choose 2}$ $K_{4}$s and $n$ $K_{3}$s
containing $ij$; hence $\eta_{ij}$ may appear at most $6{n \choose 2}+3n=3n^{2}$
times. If $ij\notin E_{b}$, then $\eta_{ij}$ appears when there
is a $K_{4}$ or a $K_{3}$ containing $ij$ and some bad edge. By
Condition 4, there are at most $2\varepsilon_{0}n$ such bad $K_{3}$s.
If the bad edge in $K_{4}$ is incident to $ij$, then there are at
most $2\varepsilon_{0}n\cdot(n-3)$ such $K_{4}$s, and if the bad
edge is not incident to $ij$, then there are at most $|E_{b}|\le\varepsilon_{0}n^{2}$
such $K_{4}$s. Thus $\eta_{ij}$ may appear at most $3\cdot2\varepsilon_{0}n+6\cdot(2\varepsilon_{0}n(n-3)+\varepsilon_{0}n^{2})\le18\varepsilon_{0}n^{2}$
times. Therefore
\begin{eqnarray*}
\sum_{ij\in E_{b}}\sum_{k\ell\in E_{g}}\eta(ij,k\ell) & \le & \sum_{ij\in E_{b}}3n^{2}\cdot\eta_{ij}+\sum_{ij\in E(K_{n})}18\varepsilon_{0}n^{2}\cdot\eta_{ij}.
\end{eqnarray*}
If $\varepsilon_0 \le \frac{c_1p^2}{8}$, then by Lemma \ref{lem:badgood}, we thus have
\[
\sum_{ij\in E_{b}}\sum_{k\ell\in E_{g}}\eta(ij,k\ell)\le\frac{24\varepsilon_{0}}{c_{1}p^{2}}n^{2}\sum_{ij\in E_{g}}\eta_{ij}+\sum_{ij\in E(K_{n})}18\varepsilon_{0}n^{2}\cdot\eta_{ij}\le\frac{42\varepsilon_{0}}{c_{1}p^{2}}n^{2}\sum_{ij\in E(K_{n})}\eta_{ij}.
\]
Hence
\[
	\frac{42\varepsilon_{0}}{c_{1}p^{2}}n^{2}\sum_{ij\in E(K_{n})}\eta_{ij}
	\ge \sum_{ij\in E_{b}}\sum_{k\ell\in E_{g}}\eta(ij,k\ell) 
	> \frac{\beta}{1024c_{0}}n^{2}p\cdot\sum_{ij\in E_{b}}|\delta_{ij}|\|t_{ij}^{(0)}\|_2.
\]
If $\varepsilon_{0}\le\frac{\beta c_{1}^{2}p^{4}}{16 \cdot 42 \cdot 1024 c_{0}}$, then
\[
	\sum_{ij\in E(K_{n})}\eta_{ij}
	> \frac{\beta c_{1}p^{3}}{42\cdot1024c_{0}\varepsilon_{0}}\sum_{ij\in E_{b}}|\delta_{ij}|\|t_{ij}^{(0)}\|_2
	> \frac{16}{c_{1}p}\sum_{ij\in E_{b}}|\delta_{ij}|\|t_{ij}^{(0)}\|_2.
\]
If $\varepsilon_0 \le \frac{c_1p^2}{8}$, then by Lemma \ref{lem:transference}, this gives 
\[
	\sum_{ij\in E_{g}}\eta_{ij}\ge\frac{c_{1}p}{8}\sum_{ij\in E(K_{n})}\eta_{ij} 
	> 2\sum_{ij\in E_{b}}|\delta_{ij}|\|t_{ij}^{(0)}\|_2.
\]
Since Lemma \ref{lem:badgood} implies $\sum_{ij\in E_{g}}\eta_{ij}\ge 2 \sum_{ij\in E_{b}}\eta_{ij}$ 
(given $\eps_0 \leq \frac{c_1 p^2}{16}$), 
together with the inequality above, we have $\sum_{ij\in E_{g}}\eta_{ij}>\sum_{ij\in E_{b}}(|\delta_{ij}|\|t_{ij}^{(0)}\|_2+\eta_{ij})$.
By (\ref{eq:gain3-3d}), this shows that $R(T)>R(T_{0})$. 
The parameters must satisfy $\varepsilon_0 \le \min \{\frac{c_1p^2}{8}, \frac{\beta c_1^2 p^4}{16 \cdot 42 \cdot 1024 c_0}, \frac{c_1p^2}{16} \}$ and $\varepsilon_1 \le \frac{p}{192c_0}$.

\medskip{}

\noindent\textbf{Case 2}. $\overline{\delta} \neq 0$ and $\sum_{ij\in E_{g}'}|\delta_{ij}|\ge\frac{1}{8}\overline{\delta}|E_{g}'|$ and $|\Eb| \neq 0$.

In this case, we first show that there are large amount of positive
and negative parallel motions on the good edges. This will imply that
there is a large amount of rotational motions on the $K_{4}$s of
the form $\{i,j,k,\ell\}$ where $ij,k\ell\in E_{g}$ and $\delta_{ij}\ge0$,
$\delta_{k\ell}<0$. Since $\|t_{ij}^{(0)}\|_2 \ge \frac{1}{2}\mu$ for all $ij \in E_g$, 
Case 2 implies 
\[
	\sum_{ij\in E_{g}}|\delta_{ij}|\|t_{ij}^{(0)}\|_2
	\ge\sum_{ij\in E_{g}'}|\delta_{ij}|\|t_{ij}^{(0)}\|_2
	\ge\frac{1}{2}\mu\cdot\sum_{ij\in E_{g}'}|\delta_{ij}|
	\ge\frac{1}{16}\mu\overline{\delta}|E_{g}'|.
\]
Define $E_{+}=\{ij\in E_{g}\,:\,\delta_{ij}\ge0\}$ and $E_{-}=\{ij\in E_{g}\,:\,\delta_{ij}<0\}$.
The inequality above and (\ref{eq:good_parallel}) implies
\begin{align*}
	\sum_{ij\in E_{+}}\delta_{ij}\|t_{ij}^{(0)}\|_2 
	&= \frac{1}{2} \sum_{ij \in \Eg} (|\delta_{ij}| + \delta_{ij}) \| \toij\| \\
	&\ge \frac{1}{2}\left(\frac{1}{16}\mu\overline{\delta}|E_{g}'| - 2\sum_{ij \in E_b} |\delta_{ij}|\|\toij\|_2 \right)
  \ge  \frac{1}{2}\left(\frac{1}{16}\mu\overline{\delta}|E_{g}'|-2c_{0}\mu\overline{\delta}|E_{b}|\right).
\end{align*}
From \eqref{eq:longgoodedges}, we have $|E_{g'}| \ge \frac{1}{16c_0}n^2p$.
Therefore if $\varepsilon_0 \le \frac{p}{1024c_0^2}$, then 
\[
	\sum_{ij\in E_{+}}\delta_{ij}\|t_{ij}^{(0)}\|_2 
	\ge \frac{1}{32}\mu\overline{\delta}\left(\frac{1}{16c_0}n^{2}p-32c_0 \eps_0 n^2\right)
	\ge\frac{1}{1024c_0}\mu\overline{\delta}n^{2}p.
\]
Similarly $\sum_{ij\in E_{-}}(-\delta_{ij})\|t_{ij}^{(0)}\|_2\ge\frac{1}{1024c_0}\mu\overline{\delta}n^{2}p$.

We either have $|E_{+}|\ge\frac{1}{2}|E_{g}|$ or $|E_{-}|>\frac{1}{2}|E_{g}|$.
If the former holds, then by Lemmas \ref{lem:rigidity1} and \ref{lem:rigidity2},
\begin{eqnarray*}
\sum_{ij\in E_{-}}\sum_{k\ell\in E_{+}}\eta(ij,k\ell) & \ge & \sum_{ij\in E_{-}}\sum_{k\ell\in E_{+}\setminus B(ij)}\frac{\beta}{4}(-\delta_{ij})\|t_{ij}^{(0)}\|_2\\
 & \ge & \frac{\beta}{4}\cdot\sum_{ij\in E_{-}}(-\delta_{ij})\|t_{ij}^{(0)}\|_2(|E_{+}|-|B(ij)|).
\end{eqnarray*}
By Lemma \ref{lem:badK4}, we have $|B(ij)|\le6\varepsilon_{1}n^{2}$,
and thus $|E_{+}|-|B(ij)|\ge\frac{1}{2}|E_{g}|-6\varepsilon_{1}n^{2}\ge\frac{1}{2}(\frac{1}{4}n^{2}p-\varepsilon_{0}n^{2})-6\varepsilon_{1}n^{2}$.
If $\varepsilon_0<\frac{1}{16}p$ and $\varepsilon_{1}\le\frac{1}{192}p$, then
$|E_{+}|-|B(ij)|\ge\frac{1}{16}n^{2}p$, and the above gives
\[
	\sum_{ij\in E_{-}}\sum_{k\ell\in E_{+}}\eta(ij,k\ell)
	\ge\frac{\beta}{4}\cdot\frac{1}{16}n^{2}p\cdot\sum_{ij\in E_{-}}(-\delta_{ij})\|t_{ij}^{(0)}\|_2
	\ge\frac{\beta}{64}\cdot\frac{1}{1024c_0}\mu\overline{\delta}n^{4}p^{2}.
\]
Similarly, if $|E_{-}|>\frac{1}{2}|E_{g}|$, then $\sum_{ij\in E_{+}}\sum_{k\ell\in E_{-}}\eta(ij,k\ell)\ge\frac{\beta}{64\cdot1024c_0}\mu\overline{\delta}n^{4}p^{2}$.

On the other hand since each edge is contained in at most $\frac{n(n-1)}{2}$
copies of $K_{4}$ and $n$ copies of $K_{3}$ (and there are 6 edges
in a $K_{4}$), we have 
\[
\sum_{ij\in E_{-}}\sum_{k\ell\in E_{+}}\eta(ij,k\ell)\le\left(6\frac{n(n-1)}{2}+3n\right)\sum_{ij\in E(K_{n})}\eta_{ij}\le3n^{2}\sum_{ij\in E(K_{n})}\eta_{ij}.
\]
If $\varepsilon_{0}\le\frac{\beta c_{1}p^{3}}{32 \cdot 3 \cdot 64 \cdot 1024 \cdot c_{0}^2}$, then
\begin{eqnarray*}
	\sum_{ij\in E(K_{n})}\eta_{ij} 
	& \ge & \frac{1}{3n^{2}}\cdot\frac{\beta}{64\cdot1024c_0}\mu\overline{\delta}n^{4}p^{2}
	=\frac{\beta p^{2}}{3\cdot64\cdot1024c_0}\mu\overline{\delta}n^{2} \\
	& > & \frac{32}{c_{1}p}c_{0}\mu\overline{\delta}|E_{b}|
	\ge\frac{32}{c_{1}p}\sum_{ij\in E_{b}}|\delta_{ij}|\|t_{ij}^{(0)}\|_2,
\end{eqnarray*}
where the last inequality follows from Condition \ref{enu:length}.
If $\varepsilon_0 \le \frac{c_1p^2}{8}$, then by Lemma \ref{lem:transference}, this implies 
\[
	\sum_{ij\in E_{g}}\eta_{ij}
	\ge\frac{c_{1}p}{16}\sum_{ij\in E(K_{n})}\eta_{ij}
	>2\sum_{ij\in E_{b}}|\delta_{ij}|\|t_{ij}^{(0)}\|_2,
\]
Therefore from (\ref{eq:gain3-3d}) and (\ref{eq:badtwist-3d}), 
\begin{eqnarray*}
	R(T) & \ge & R(T_{0})+\sum_{ij\in E_{g}}\eta_{ij}-\sum_{ij\in E_{b}}(|\delta_{ij}|\|t_{ij}^{(0)}\|_2+\eta_{ij})\\
	& > & R(T_{0})+2\sum_{ij\in E_{b}}|\delta_{ij}|\|t_{ij}^{(0)}\|_2-\sum_{ij\in E_{b}}2|\delta_{ij}|\|t_{ij}^{(0)}\|_2=R(T_{0}).
\end{eqnarray*}
The parameters must satisfy $\varepsilon_0 \le \min\{\frac{p}{1024c_0^2}, \frac{1}{16}p, \frac{\beta c_1 p^3}{32 \cdot 3 \cdot 64 \cdot 1024 \cdot c_0^2}, \frac{c_1p^2}{8}\}$ and $\varepsilon_1 \le \frac{1}{192}p$.
\end{proof}

\subsection{Properties of Gaussians in three dimensions} \label{sec-gaussians-well-distributed-3d}

The first lemma establishes a bound on the average distance between random
Gaussian vectors.

\begin{lemma} \label{lem:gaussian_length}
There exists a positive constant $c$ such that if $G$ is a $p$-typical graph with vertex set $[n]$, then with probability at least 
$1 - 3ne^{-cnp/2}$, 
\[
	\sum_{ij \in E(G)} \|t_i - t_j\|_2 \ge \frac{1}{8} n^2p.
\]
\end{lemma}
\begin{proof}
%
%
Let $v \in \mathbb{R}^3$ be a fixed vector. 
Note that for all $j \in [n]$, we have
\[
	\|v - t_j\|_2^2
	= \|v\|^2 + \|t_j\|^2 - 2 \langle v, t_j \rangle.
\]
Therefore if $\langle v, t_j \rangle \le 0$, then $\|v - t_j\|_2 \ge \|t_j\|_2$. 
Further, by the symmetry of Gaussian random variables, we know that
the distribution of $\|t_j\|_2$ remains the same even after conditioning on the event
$\langle v, t_j \rangle \le 0$. Therefore
\[
	\mathbb{E}[\|v - t_j\|_2]
	\ge \P\Big(\langle v,t_j\rangle \le 0\Big) \cdot \mathbb{E}\Big[\|t_j\|_2 \,\Big|\, \langle v, t_j \rangle \le 0\Big]
	= \frac{1}{2} \mathbb{E}[\|t_j\|_2]
	= \sqrt{\frac{2}{\pi}},
\]
where the final equality holds since each $\|t_j\|_2$ is subgaussian with mean $\sqrt{8/\pi}$.  
Fix an index $i \in [n]$ and let $N_i$ be the neighborhood of $i$ in $G$. 
Since $G$ is $p$-typical, we have $|N_i| \ge \frac{1}{2}np$. 
By the analysis above, we see that
\[
	\mathbb{E}\left[ \sum_{j \in N_i} \|t_i - t_j\|_2 \right]
	\ge |N_i|\sqrt{\frac{2}{\pi}}.
\]
By Proposition 5.10 in Vershynin \cite{Vershynin} on the concentration of subgaussians, there is a constant $c$ such that
with probability at least $1 - e^{1-c|N_i|}$, we have
$\sum_{j \in N_i} \|t_i - t_j\|_2 \ge \frac{1}{2}|N_i| \ge \frac{1}{4}np$.
Therefore by taking the union bound over all indices $i \in [n]$, 
we see that with probability at least $1 - 3ne^{-cnp/2}$, 
\[
	\sum_{ij \in E(G)} \|t_i - t_j\|_2
	\ge \frac{1}{2} \sum_{i \in [n]}\sum_{j \in N_i} \|t_i - t_j\|_2
	\ge \frac{n}{2} \cdot \frac{1}{4}np = \frac{1}{8}n^2p. \qedhere
\]
\end{proof}

The second lemma establishes a bound on the angle between random Gaussian vectors.

\begin{lemma} \label{lem:gaussian_angle}
Let $x,y\in\mathbb{R}^{3}$ be linearly independent vectors.
If $t_{1},t_{2},\cdots,t_{n}\in\mathbb{R}^{3}$ are independent random
Gaussian vecotrs, then with probability $1-e^{-\Omega(\beta n)}$,
for all but at most $\beta n$ vectors $t_{i}$, we have 
\[
1-\left\langle \frac{t_{i}-x}{\|t_{i}-x\|_2},\,\frac{y-x}{\|y-x\|_2}\right\rangle ^{2}\ge\frac{\beta^{2}}{2(\|t_{i}\|_2^{2}+\|x\|_2^{2})}.
\]
\end{lemma}
\begin{proof}
Fix an index $i\in[n]$. Note that 
\begin{eqnarray}
1-\left\langle \frac{t_{i}-x}{\|t_{i}-x\|_2},\,\frac{y-x}{\|y-x\|_2}\right\rangle ^{2} & = & \left\Vert \proj_{(y-x)^{\perp}}\frac{t_{i}-x}{\|t_{i}-x\|_2}\right\Vert_2^{2}=\frac{\|\proj_{(y-x)^{\perp}}(t_{i}-x)\|_2^{2}}{\|t_{i}-x\|_2^{2}} \nonumber\\
 & \ge & \frac{\|\proj_{\{x,y\}^{\perp}}t_{i}\|_2^{2}}{\|t_{i}-x\|_2^{2}}\ge\frac{\|\proj_{\{x,y\}^{\perp}}t_{i}\|_2^{2}}{2(\|t_{i}\|_2^{2}+\|x\|_2^{2})}. \label{eq:ga_eq1}
\end{eqnarray}
Since $t_{i}$ is a random Gaussian vector and $x,y$ are linearly independent, 
the distribution of $\|\proj_{\{x,y\}^{\perp}}t_{i}\|_2$
is that of the absolute value of a standard normal distribution. Therefore
$\P(\|\proj_{\{x,y\}^{\perp}}t_{i}\|_2<\beta)\le\frac{1}{\sqrt{2\pi}}\int_{-\beta}^{\beta}e^{-x^{2}/2}dx\le\sqrt{\frac{2}{\pi}}\beta$.
Let ${\bf 1}_i$ be the indicator random variable of the event that
$\|\proj_{\{x,y\}^{\perp}}t_{i}\|_2<\beta$. We seen above that $\mathbb{E}[{\bf 1}_i] < \sqrt{\frac{2}{\pi}}\beta$.
Further, since $\{t_{i}\}_{i \in [n]}$ are independent, it follows that $\{{\bf 1}_i\}_{i \in [n]}$ are independent.
Therefore by Chernoff's inequality,
\[
	\P\left( \sum_{i \in [n]} {\bf 1}_i - \sqrt{\frac{2}{\pi}}\beta n > \frac{1}{5} \beta n \right)
	\le e^{-\Omega(\beta n)}.
\]
Hence with probability $1-e^{-\Omega(\beta n)}$,
there are at most $\sqrt{\frac{2}{\pi}}\beta n + \frac{1}{5} \beta n < \beta n$ vectors $t_{i}$ for which $\|\proj_{\{x,y\}^{\perp}}t_{i}\|_2<\beta$. The lemma now follows from \eqref{eq:ga_eq1}.
\end{proof}

The next lemma shows that random Gaussian vectors are well-distributed
with respect to a fixed pair of vectors.

\begin{lemma} \label{lem:gaussian_wd_prim}
There exists a positive real number $c$
such that the following holds for all pairs of linearly independent vectors $x,y\in\mathbb{R}^{3}$.
If $t_{1},\cdots,t_{n}\in\mathbb{R}^{3}$ are independent random Gaussian
vectors, then with probability $1-e^{-\Omega(n)}$,
the set of vectors $\{t_{1},\cdots,t_{n}\}$ are $\frac{c}{\max\{1,\|x+y\|_2\}}$-well-distributed
with respect to $(x,y)$. 
\end{lemma}
\begin{proof}
Let $c$ be a positive real number to be chosen later. We may rotate
the vectors so that $x=(\ell,0,x_{3})$ and $y=(\ell,0,y_{3})$ for
some $x_{3},y_{3},\ell\in\mathbb{R}$ where $\ell\ge0$. Note that
$\|x+y\|_2\ge2\ell$. Define $\ell_{0}=\max\{1,\ell\}$. It suffices
to give an estimate on the probability that
\[
\sum_{i=1}^{n}\|\proj_{span\{t_{i}-x,t_{i}-y\}^{\perp}}(h)\|_2\ge\frac{cn}{\ell_{0}}
\]
holds for all vectors $h\in(x-y)^{\perp}=\{(a,b,0):a,b\in\mathbb{R}\}$
satisfying $\|h\|_2=1$. 

Fix a vector $h=(h_{1},h_{2},0)$ satisfying $\|h\|_2=1$. For $t_{i}=(t_{i,1},t_{i,2},t_{i,3})$,
we have $span\{t_{i}-x,t_{i}-y\}=span\{(0,0,1),x+y-2t_{i}\}=span\{(0,0,1),(2\ell-2t_{i,1},-2t_{i,2},0)\}$.
Hence $s=(t_{i,2},\ell-t_{i,1},0)\in span\{t_{i}-x,t_{i}-y\}^{\perp}$, and
\begin{eqnarray*}
	\|\proj_{span\{t_{i}-x,t_{i}-y\}^{\perp}}(h)\|_2 
	& = & \frac{|\langle s,h\rangle |}{\|s\|_2}=\left|\frac{(t_{i,2},\ell-t_{i,1},0)\cdot h}{\sqrt{t_{i,2}^{2}+(\ell-t_{i,1})^{2}}}\right|\\
 & = & \frac{\left|h_{1}t_{i,2}+(\ell-t_{i,1})h_{2}\right|}{\sqrt{t_{i,2}^{2}+(\ell-t_{i,1})^{2}}}.
\end{eqnarray*}
Assume that $h_{1}\ge h_{2}\ge0$, which implies $h_{1}\ge\frac{1}{\sqrt{2}}$.
Since $t_{i,1}$ is normally distributed with variance 1, the probability
that $-1\le t_{i,1}\le0$ is $p$ for some fixed postive real number
$p$. Conditioned on this event and the event that $t_{i,2}\ge0$
(note that $t_{i,1}$ and $t_{i,2}$ are independent), we have 
\begin{eqnarray*}
	\|\proj_{span\{t_{i}-x,t_{i}-y\}^{\perp}}(h)\|_2
	& = & \frac{\left|h_{1}t_{i,2}+(\ell-t_{i,1})h_{2}\right|}{\sqrt{t_{i,2}^{2}+(\ell-t_{i,1})^{2}}}\ge\frac{h_{1}t_{i,2}}{\sqrt{t_{i,2}^{2}+4\ell_{0}^{2}}}.
\end{eqnarray*}
Therefore 
\[
	\P\left(\|\proj_{span\{t_{i}-x,t_{i}-y\}^{\perp}}(h)\|_2
	>\frac{1}{2\ell_{0}}\right)\ge\P\left(\frac{h_{1}t_{i,2}}{\sqrt{t_{i,2}^{2}+4\ell_{0}^{2}}}>\frac{1}{2\ell_{0}}\,\Big|\, t_{i,2}\ge0\right)\cdot\frac{1}{2}p.
\]
Note that for $t_{i,2}\ge0$, the inequality $\frac{h_{1}t_{i,2}}{\sqrt{t_{i,2}^{2}+4\ell_{0}^{2}}}>\frac{1}{2\ell_{0}}$
is equivalent to $h_{1}^{2}t_{i,2}^{2}>\frac{t_{i,2}^{2}}{4\ell_{0}^{2}}+1$,
which is equivalent to $t_{i,2}^{2}(h_{1}^{2}-\frac{1}{4\ell_{0}^{2}})>1$.
Since $h_{1}^{2}\ge\frac{1}{2}$ and $\ell_{0}\ge1$, we have 
\[
	\P\left(\|\proj_{span\{t_{i}-x,t_{i}-y\}^{\perp}}(h)\|_2>\frac{1}{2\ell_{0}}\right)
	\ge\P(t_{i,2}^{2}>4\,|\, t_{i,2}\ge0)\cdot\frac{1}{2}p=q
\]
for some fixed positive real number $q$. 
By considering the indicator random variable of the events 
$\|\proj_{span\{t_{i}-x,t_{i}-y\}^{\perp}}(h)\|_2>\frac{1}{2\ell_{0}}$, we see by Chernoff's inequality
that with probability $1-e^{-\Omega(n)}$, there are at least $\frac{qn}{2}$ indices 
$i \in [n]$ such that $\|\proj_{span\{t_{i}-x,t_{i}-y\}^{\perp}}(h)\|_2>\frac{1}{2\ell_{0}}$.
Note that this implies
\begin{eqnarray*}
	\sum_{i=1}^{n}\|\proj_{span\{t_{i}-x,t_{i}-y\}^{\perp}}(h)\|_2 
	\ge \frac{qn}{2} \cdot \frac{1}{2\ell_0}
	= \frac{qn}{4\ell_{0}}.
\end{eqnarray*}
To handle the case of $h_2 \geq h_1 \geq 0$, note that if $t_{i,1} \le 0$ and $0 \le t_{i,2} \le 1$, then
\begin{eqnarray*}
	\|\proj_{span\{t_{i}-x,t_{i}-y\}^{\perp}}(h)\|_2
	& = & \frac{\left|h_{1}t_{i,2}+(\ell-t_{i,1})h_{2}\right|}{\sqrt{t_{i,2}^{2}+(\ell-t_{i,1})^{2}}}
	\ge \frac{(\ell-t_{i,1})h_{2}}{\sqrt{1+(\ell-t_{i,1})^{2}}}
	\ge \frac{1}{\sqrt{2}} \cdot \frac{\ell-t_{i,1}}{\sqrt{1+(\ell-t_{i,1})^{2}}}.
\end{eqnarray*}
Since $\frac{x}{\sqrt{1+x^2}}$ is decreasing in the range $x \ge 0$, 
if $t_{i,1} \le -1$, then $\|\proj_{span\{t_{i}-x,t_{i}-y\}^{\perp}}(h)\|_2 \ge \frac{1}{2}$. 
Therefore we see as in above that with probability $1-e^{-\Omega(n)}$, there are at least $\frac{qn}{2}$ indices 
$i \in [n]$ such that $\|\proj_{span\{t_{i}-x,t_{i}-y\}^{\perp}}(h)\|_2>\frac{1}{2} \ge \frac{1}{2\ell_0}$.
All the remaining cases can be handled analogously.

Let $H$ be a set of $\lceil 2\pi \cdot \frac{8\ell_{0}}{q} \rceil \le \frac{60\ell_0}{q}$
vectors uniformly distributed along the circle $S_2 = \{(x,y,0) \,:\, x^2 + y^2 = 1\}$. 
Apply the analysis above to
each vector in $H$ and take the union bound to conclude that with
probability $1-\ell_{0}e^{-\Omega(n)}$, for all $h\in H$,
\[
\sum_{i=1}^{n}\|\proj_{span\{t_{i}-x,t_{i}-y\}^{\perp}}(h)\|_2\ge\frac{q}{4}\frac{n}{\ell_{0}}.
\]
Let $h' \in S_2$ be an arbitrary vector and
let $h \in H$ be the vector closest to $h'$. The distance from $h$ to $h'$ along the circle $S_2$ is at most 
$2\pi \cdot \frac{1}{|H|} \le \frac{q}{8\ell_0}$, and hence $\|h - h'\|_2 \le \frac{q}{8\ell_0}$. 
Thus for all $i$, we have 
\begin{eqnarray*}
\|\proj_{span\{t_{i}-x,t_{i}-y\}^{\perp}}(h')\|_2 & \ge & \|\proj_{span\{t_{i}-x,t_{i}-y\}^{\perp}}(h)\|_2-\|h-h'\|_2\\
 & \ge & \|\proj_{span\{t_{i}-x,t_{i}-y\}^{\perp}}(h)\|_2-\frac{q}{8\ell_{0}}
\end{eqnarray*}
Therefore 
\[
\sum_{i=1}^{n}\|\proj_{span\{t_{i}-x,t_{i}-y\}^{\perp}}(h')\|_2\ge\frac{q}{4}\frac{n}{\ell_{0}}-\frac{q}{8}\frac{n}{\ell_{0}}\ge\frac{q}{8}\frac{n}{\ell_{0}}.
\]
Since $\ell_{0}=\max\{1,\ell\}\le\max\{1,\|x+y\|_2\}$, this implies
the lemma for $c=\frac{q}{8}$.
\end{proof}

By applying the union bound together with the three lemmas above,
we obtain the following lemma.

\begin{lemma} \label{lem:gaussian_wd}
There exists $c, \zeta \in \mathbb{R}$ and $n_0 \in \mathbb{N}$ such
that the following holds for all positive real numbers $\varepsilon$ and natural numbers $n \ge n_0$.
Let $G$ be a $p$-typical graph with vertex set $[n]$ for some $p$ satisfying 
$np^2 \geq \zeta \log n$. If $t_{1},\cdots,t_{n}\in\mathbb{R}^{3}$
are independent random Gaussian vectors, then the following holds
with probability $1-n^{-5}$,
\begin{itemize}
  \setlength{\itemsep}{1pt} \setlength{\parskip}{0pt}
  \setlength{\parsep}{0pt}
\item[1.] For each distinct $i,j \in [n]$, for all but at most $\varepsilon n$ indices $k \in [n]$, we have $1 - \langle \frac{t_k - t_i}{\|t_k - t_i \|_2}, \frac{t_j - t_i}{\|t_j - t_i\|_2} \rangle^2 \ge \frac{\varepsilon^2}{64 \log n}$,
\item[2.] for all distinct $i,j \in [n]$, we have $\|t_i - t_j\|_2 \le 40\sqrt{\log n} \cdot \mu$, where $\mu = \frac{1}{|E(G)|} \sum_{ij \in E(G)} \|t_i - t_j\|_2$, and
\item[3.] the set $\{t_i\}_{i\in[n]}$ is $\frac{c}{\sqrt{\log n}}$-well-distributed along $G$.
\end{itemize}\end{lemma}
\begin{proof}
Let $c$ be eight times the constant coming from Lemma \ref{lem:gaussian_wd_prim}.
For each distinct $i,j\in[n]$, define $S_{ij}=\{t_{k}\,:\, ik,jk\in E(G)\}$.
Consider the following events:
\begin{itemize}
  \setlength{\itemsep}{1pt} \setlength{\parskip}{0pt}
  \setlength{\parsep}{0pt}
\item[(i)] for all $i\in[n]$, we have $\|t_i\|_2 \le  4\sqrt{\log n}$,
\item[(ii)] $\sum_{ij \in E(G)} \|t_i - t_j \|_2 \ge \frac{1}{8}n^2p$, 
\item[(iii)] for each distinct $i,j \in [n]$, for all but at most $\varepsilon n$ integers $k \in [n]$, we have $1 - \langle \frac{t_k - t_i}{\|t_k - t_i \|_2}, \frac{t_j - t_i}{\|t_j - t_i\|_2} \rangle^2 \ge \frac{\varepsilon^2}{2(\|t_k\|_2^2 + \|t_i\|_2^2)}$,
\item[(iv)] for each distinct $i,j\in[n]$, $S_{ij}$ is $\frac{8c}{\max\{1,\|t_i+t_j\|_2\}}$-well-distributed with respect to $(t_i,t_j)$.
\end{itemize}
For a fixed $i \in [n]$, since $\|t_i\|_2^2$ follows a $\chi^2$ distribution with $3$ degrees of freedom, 
standard estimates on Chi-squared random variables, such as Lemma 1 in \cite{LaurentMassart}, give
\[
\P( \|t_i\|_2^2 \geq 3 + 2 \sqrt{3}t + 2t^2 ) \leq e^{-t^2}.
\]
Let $t = \sqrt{7 \log n}$. 
If $n$ is sufficiently large, then 
$2t^2 + 2 \sqrt{3} t + 3 < 16 \log n$. 
As a result, $\P(\|t_i\|_2^2 \geq 16 \log n) \leq e^{-7 \log n}$.
Hence
Property (i) holds with probability $1-n^{-6}$ by taking the union
bound over all $i \in [n]$. 
Property (ii) holds with probability $1-e^{-\Omega(n)}$ by
Lemma \ref{lem:gaussian_length}. For a fixed pair $i,j\in[n]$, Property
(iii) holds with probability $1-e^{-\Omega(\varepsilon n)}$ by Lemma \ref{lem:gaussian_angle}.
Hence by taking the union bound, we see that Property (iii) holds
with probability $1-n^{2}e^{-\Omega(\varepsilon n)}$. For a fixed pair $i,j\in[n]$,
by Lemma~\ref{lem:gaussian_wd_prim}
and the fact that each pair is contained in at least $\frac{1}{2}np^2$ triangles, 
we have Property (iv) for the pair $i,j$
with probability $1-e^{-\Omega(np^2)}$. Hence by taking
the union bound, we see that Property (iv) holds with probability
$1-n^{2}e^{-\Omega(np^2)}$. 
Thus we see that all four
events (i)-(iv) simultaneously hold with at least probability $1-n^{-5}$ for sufficiently large $n$, provided that $np^2 \geq \zeta \log n$ for sufficiently large $\zeta$.

We now show that Properties (i)-(iv) imply Properties 1-3. Note that
Properties 1 and 3 immediately follow from Properties (i), (iii), and
(iv). Further, since $|E(G)|\le n^2p$, Property (ii) implies
\begin{eqnarray*}
	\mu & = & \frac{1}{|E(G)|}\sum_{ij\in E(G)}\|t_{i}-t_{j}\|_2
	\ge\frac{1}{n^2p} \cdot \frac{1}{8}n^2 p = \frac{1}{8}.
\end{eqnarray*}
Hence by Property (i), we have for all $i,j\in[n]$, 
\[
	\|t_{i}-t_{j}\|_2
	\le \|t_{i}\|_2+\|t_{j}\|_2
	\le 8\sqrt{\log n}\le 64\mu \sqrt{\log n}.\qedhere
\]
\end{proof}

\subsection{Proof of Theorem \ref{thm-3d}} \label{sec-proof-random-theorem-3d}

We can now prove the three-dimensional recovery theorem, which we state here again for convenience:
\setcounter{theorem}{1}
\begin{theorem}
There exists $n_0 \in \mathbb{N}$ and $c \in \mathbb{R}$ such that the following holds for all $n \ge n_0$.
Let $G([n],E)$ be drawn from $G(n,p)$ for some $p = \Omega( n^{-1/5} \log^{3/5} n)$. Take $ \tnot_1, \ldots \tnot_n  \in \R^3$, where $\toi \sim \mathcal{N}(0, I_{3 \times 3})$ are i.i.d., independent from $G$. 
There exists $\gamma = \Omega(p^5 / \log^3 n)$ and an event of probability at least  $1- \frac{1}{n^{4}}$ 
on which the following holds:\\[1em]
For arbitrary  subgraphs $\Eb$ satisfying $\max_i \deg_b(i) \leq \gamma n$ and arbitrary pairwise direction corruptions $\vij \in \mathbb{S}^2$ for $ij \in \Eb$,  the convex program \eqref{shapefit} has a unique minimizer equal to $\left \{\alpha \Bigl(\toi - \tnotbar \Bigr)\right\}_{i \in [n]}$ for some positive $\alpha$ and for $\tnotbar = \frac{1}{n}\sum_{i\in [n]} \toi$. 
\end{theorem}

\begin{proof}
Let $n_0$ be a sufficiently large natural number larger than
that coming from Lemma~\ref{lem:gaussian_wd}.
Lemma~\ref{lem:ptyp}  implies $G$ is $p$-typical with probability
$1 - n^2 e^{-\Omega(np^2)}$. Condition on $G$ being $p$-typical.
Let $c$ be the constant from Lemma \ref{lem:gaussian_wd}.
By applying Lemma \ref{lem:gaussian_wd}
with $\varepsilon=\frac{p}{2^{15}\sqrt{\log n}}$, with probability
at least $1-n^{-5}$, we have
\begin{itemize}
  \setlength{\itemsep}{1pt} \setlength{\parskip}{0pt}
  \setlength{\parsep}{0pt}
\item[1.] For each distinct $i,j \in [n]$ satisfying $i < j$, for all but at most $2 \eps n = \frac{p}{2^{14}\sqrt{\log n}}$ integers $k \in [n]$, we have $1 - \langle \frac{t_k - t_i}{\|t_k - t_i \|}, \frac{t_j - t_i}{\|t_j - t_i\|} \rangle^2 \ge \frac{p^2}{2^{36} \log^2 n}$ and $1 - \langle \frac{t_k - t_j}{\|t_k - t_j \|}, \frac{t_i - t_j}{\|t_i - t_j\|} \rangle^2 \ge \frac{p^2}{2^{36} \log^2 n}$,
\item[2.] for all distinct $i,j \in [n]$, we have $\|t_i - t_j\| \le 64\sqrt{\log n} \cdot \mu$, where $\mu = \frac{1}{|E(G)|} \sum_{ij \in E(G)} \|t_i - t_j\|$, and
\item[3.] the set $\{t_i\}_{i\in[n]}$ is $\frac{c}{\sqrt{\log n}}$-well-distributed along $G$.
\end{itemize}
Thus the probability that $G$ is $p$-typical and Properties 1-3 listed above holds
is at least $1 - n^{-4}$. 
Hence we may apply Theorem \ref{thm-deterministic-three-d} with with 
$c_{0}=64\sqrt{\log n}$,
$\varepsilon_{1}= \frac{p}{2^{14}\sqrt{\log n}} \le \frac{p}{192c_{0}}$,
$\beta=\sqrt{\frac{p^2}{2^{36}\log^{2}n}}=\frac{p}{2^{18}\log n}$,
and $c_{1}=\frac{c}{\sqrt{\log n}}$. The theorem holds if 
\[
	\varepsilon_{0}
	\le \frac{c^2p^5}{2^{53} \log^{3}n}
	\le  \frac{p}{2^{18}\log n} \cdot \frac{c^2}{\log n} p^4 \cdot \frac{1}{32 \cdot 3 \cdot 64 \cdot 1024 \cdot 64^2 \log n}
		= \frac{\beta c_1^2 p^4}{32 \cdot 3 \cdot 64 \cdot 1024\  c_0^2}.		
\]
Letting $\gamma$ from the theorem statement be $\eps_0$, note that the condition $\max_i \deg_b(i) \leq \gamma n$ is nontrivial when $p = \Omega(n^{-1/5} \log^{3/5} n)$.  \qedhere
\end{proof}

\section{Numerical simulations} \label{sec-simulations}

In this section, we use numerical simulation to verify that ShapeFit recovers locations in $\R^3$ in the presence of corrupted pairwise direction measurements.  Further, we empirically demonstrate that ShapeFit is robust to noise in the uncorrupted measurements.

Let the graph of observations be an \erdosrenyi graph $G(n,p)$ for $p = 1/2$.  Let $\ttildeoi \in \R^3$ be independent $\mathcal{N}(0, I_{3\times 3})$ random variables for $i = 1 \ldots n$.   Let $\toi = \ttildeoi - \frac{1}{n} \sum_j \ttildeoj$.
  For $ij \in E(G)$, let
\[
\vtildeij = \begin{cases} \zij & \text{ with probability } q  \\[0.2em]
\frac{\toi - \toj}{\| \toi - \toj\|_2} + \sigma \zij & \text{ with probability } 1-q
\end{cases}
\]
where $\zij$ are independent and uniform over $\mathbb{S}^2$. Let $\vij = \vtildeij / \| \vtildeij\|_2$. That is, each observation is corrupted with probability $q$, and each corruption is in a random direction.  In the noiseless case, with $\sigma = 0$, each observation is exact with probability $1-q$.  

We solved ShapeFit using the SDPT3 solver \cite{TTT1999, TTT2003} and YALMIP \cite{L2004}.  For output $T = \{ t_i\}_{i \in [n]}$, define its relative error with respect to $\Tnot = \{\toi\}_{i \in [n]}$ as
\[
\left \| \frac{T}{\|T\|_F} - \frac{\Tnot}{\|\Tnot\|_F} \right \|_F
\]
where $\|T\|_F$ is the Frobenius norm of the matrix whose column are $\{t_i\}$.    This error metric amounts to an $\ell_2$ norm after rescaling.

\begin{figure} 
\begin{center}
{\large
%
%
\begin{psfrags}%
\psfragscanon%
%
\psfrag{s01}[][]{\color[rgb]{0,0,0}\setlength{\tabcolsep}{0pt}\begin{tabular}{c}ShapeFit under corruptions and noise\end{tabular}}%
\psfrag{s10}[][]{\color[rgb]{0,0,0}\setlength{\tabcolsep}{0pt}\begin{tabular}{c}Number of locations ($n$)\end{tabular}}%
\psfrag{s11}[][]{\color[rgb]{0,0,0}\setlength{\tabcolsep}{0pt}\begin{tabular}{c}Corruption probability ($q$)\end{tabular}}%
\psfrag{s12}[][]{\color[rgb]{0,0,0}\setlength{\tabcolsep}{0pt}\begin{tabular}{c}ShapeFit under corruptions and no noise\end{tabular}}%
\psfrag{s13}[][]{\color[rgb]{0,0,0}\setlength{\tabcolsep}{0pt}\begin{tabular}{c}Number of locations ($n$)\end{tabular}}%
\psfrag{s14}[][]{\color[rgb]{0,0,0}\setlength{\tabcolsep}{0pt}\begin{tabular}{c}Corruption probability ($q$)\end{tabular}}%
%
\color[rgb]{0.15,0.15,0.15}%
%
\psfrag{x01}[t][t]{0}%
\psfrag{x02}[t][t]{0.1}%
\psfrag{x03}[t][t]{0.2}%
\psfrag{x04}[t][t]{0.3}%
\psfrag{x05}[t][t]{0.4}%
\psfrag{x06}[t][t]{0.5}%
\psfrag{x07}[t][t]{0}%
\psfrag{x08}[t][t]{0.1}%
\psfrag{x09}[t][t]{0.2}%
\psfrag{x10}[t][t]{0.3}%
\psfrag{x11}[t][t]{0.4}%
\psfrag{x12}[t][t]{0.5}%
%
\psfrag{v01}[r][r]{80}%
\psfrag{v02}[r][r]{60}%
\psfrag{v03}[r][r]{40}%
\psfrag{v04}[r][r]{20}%
\psfrag{v05}[r][r]{80}%
\psfrag{v06}[r][r]{60}%
\psfrag{v07}[r][r]{40}%
\psfrag{v08}[r][r]{20}%
%
\resizebox{14cm}{!}{\includegraphics{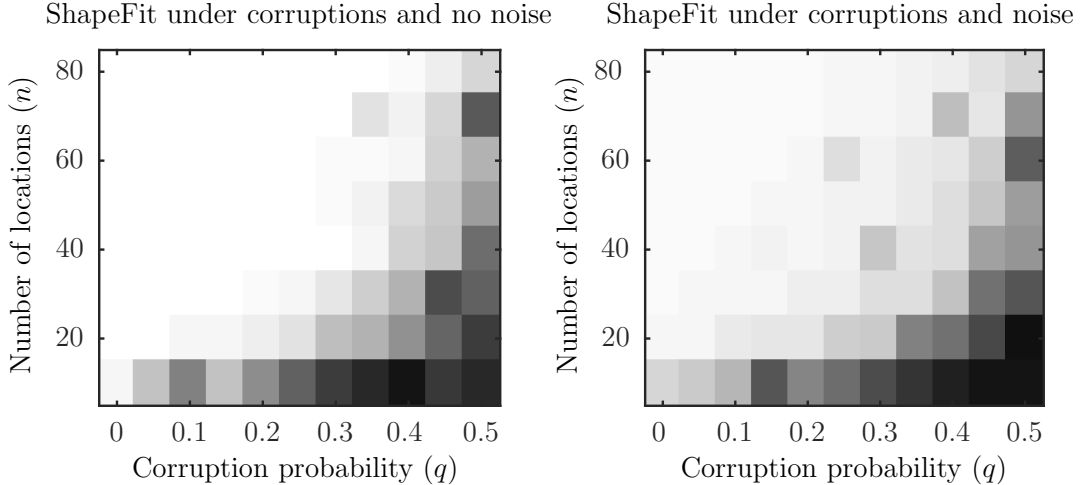}}%
\end{psfrags}%
%

}
\end{center}
\caption{Average recovery error of ShapeFit as a function of the number of locations $n$ and the corruption probability $q$.  The data model has $n$ 3d Gaussian locations whose pairwise directions are observed in accordance with an \erdosrenyi graph $G(n, 1/2)$ and are corrupted with probability $q$.   White blocks represent an average recovery error of zero over 10 independently generated problems.  Black blocks represent an average recovery error of 100\%.   The left panel corresponds to the noiseless case $\sigma = 0$, and the right panel corresponds to the noisy case $\sigma = 0.05$.  
}
\label{fig-phase-transition}
\end{figure}

Figure \ref{fig-phase-transition} shows the average residual of the output of ShapeFit over 10 independent trials for locations in $\R^3$ generated by $p = 1/2$, $\sigma \in \{0, 0.05\}$, and a range of values $10 \leq n \leq 80$ and $0 \leq q \leq 0.5$.  White blocks represent zero average residual, and black blocks represent an average residual of 1 or higher.  Average residuals between $0$ and $1$ are represented by the appropriate shade of gray.  The figure shows that ShapeFit successfully recovers 3d locations in the presence of a surprisingly large probability of corruption, provided $n$ is big enough. For example, if $n \geq 50$, recovery succeeds even when around 25\% of all measurements are randomly corrupted.  Further, successful recovery occurs both in the noiseless case, and in the noisy case with $\sigma=0.05$.

\begin{figure}
\begin{center}
{\large
%
%
\begin{psfrags}%
\psfragscanon%
%
\psfrag{s01}[][]{\color[rgb]{0,0,0}\setlength{\tabcolsep}{0pt}\begin{tabular}{c}Average residual\end{tabular}}%
\psfrag{s03}[][]{\color[rgb]{0,0,0}\setlength{\tabcolsep}{0pt}\begin{tabular}{c}Noise parameter $\sigma$\end{tabular}}%
\psfrag{s04}[][]{\color[rgb]{0,0,0}\setlength{\tabcolsep}{0pt}\begin{tabular}{c}ShapeFit for $n=50$, $q=0.2$\end{tabular}}%
%
\color[rgb]{0.15,0.15,0.15}%
%
\psfrag{x01}[t][t]{$10^{-6}$}%
\psfrag{x02}[t][t]{$10^{-5}$}%
\psfrag{x03}[t][t]{$10^{-4}$}%
\psfrag{x04}[t][t]{$10^{-3}$}%
\psfrag{x05}[t][t]{$10^{-2}$}%
\psfrag{x06}[t][t]{$10^{-1}$}%
\psfrag{x07}[t][t]{$10^{0}$}%
%
\psfrag{v01}[r][r]{$10^{-6} \hspace{-0.1em}$}%
\psfrag{v02}[r][r]{}%
\psfrag{v03}[r][r]{$10^{-4}$}%
\psfrag{v04}[r][r]{}%
\psfrag{v05}[r][r]{$10^{-2}$}%
\psfrag{v06}[r][r]{}%
\psfrag{v07}[r][r]{$10^{0}\ \hspace{0.1825em} $}%
%
\resizebox{9cm}{!}{\includegraphics{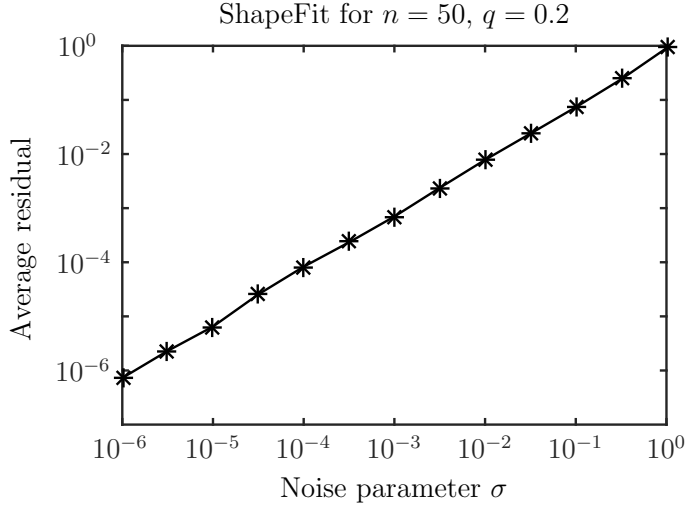}}%
\end{psfrags}%
%

}
\end{center}
\caption{
Average recovery error of ShapeFit versus the noise parameter $\sigma$.  These simulations are based on $n=50$  Gaussian locations in $\R^3$ whose pairwise directions are observed in accordance with an \erdosrenyi graph $G(n,1/2)$ and are corrupted with probability $q=0.2$.  The average is based on 10 independently generated problems.
}
 \label{fig-robustness}
\end{figure}

Figure \ref{fig-robustness} shows the average residual over 10 independent trials for locations in $\R^3$ generated by $p = 1/2$, $n = 40$, $q= 0.2$ and a range of values of $10^{-6} \leq \sigma \leq 10^0$.  We see that ShapeFit is empirically stable to noise, with average residuals that are approximately linear in the noise parameter $\sigma$.
 
\section*{Acknowledgements}
VV acknowledges discussions with Tomasz Malisiewicz, Stefano Soatto, and Ram Sripracha. VV is partially supported by the Office of Naval Research.  CL is partially supported by the National Science Foundation Grant DMS-1362326. PH is partially supported by the National Science Foundation Grant DMS-1418971.
\bibliographystyle{plain}
\bibliography{references}

\begin{thebibliography}{10}

\bibitem{AngluinValiant}
Dana Angluin and Leslie~G. Valiant.
\newblock Fast probabilistic algorithms for hamiltonian circuits and matchings.
\newblock In {\em Proceedings of the Ninth Annual ACM Symposium on Theory of
  Computing}, STOC '77, pages 30--41, New York, NY, USA, 1977. ACM.

\bibitem{1dSfm_3}
Mica Arie-Nachimson, Shahar~Z Kovalsky, Ira Kemelmacher-Shlizerman, Amit
  Singer, and Ronen Basri.
\newblock Global motion estimation from point matches.
\newblock In {\em 3D Imaging, Modeling, Processing, Visualization and
  Transmission (3DIMPVT), 2012 Second International Conference on}, pages
  81--88. IEEE, 2012.

\bibitem{1dSfm_4}
Matthew Brand, Matthew Antone, and Seth Teller.
\newblock Spectral solution of large-scale extrinsic camera calibration as a
  graph embedding problem.
\newblock In {\em Computer Vision-ECCV 2004}, pages 262--273. Springer, 2004.

\bibitem{1dSfm_5}
Avishek Chatterjee and Venu~Madhav Govindu.
\newblock Efficient and robust large-scale rotation averaging.
\newblock In {\em Computer Vision (ICCV), 2013 IEEE International Conference
  on}, pages 521--528. IEEE, 2013.

\bibitem{1dSfm_6}
David Crandall, Andrew Owens, Noah Snavely, and Dan Huttenlocher.
\newblock Discrete-continuous optimization for large-scale structure from
  motion.
\newblock In {\em Computer Vision and Pattern Recognition (CVPR), 2011 IEEE
  Conference on}, pages 3001--3008. IEEE, 2011.

\bibitem{1dSfm_7}
Peter Eades, Xuemin Lin, and William~F Smyth.
\newblock A fast and effective heuristic for the feedback arc set problem.
\newblock {\em Information Processing Letters}, 47(6):319--323, 1993.

\bibitem{1dSfm_8}
Olof Enqvist, Fredrik Kahl, and Carl Olsson.
\newblock Non-sequential structure from motion.
\newblock In {\em Computer Vision Workshops (ICCV Workshops), 2011 IEEE
  International Conference on}, pages 264--271. IEEE, 2011.

\bibitem{1dSfm_10}
Johan Fredriksson and Carl Olsson.
\newblock Simultaneous multiple rotation averaging using lagrangian duality.
\newblock In {\em Computer Vision--ACCV 2012}, pages 245--258. Springer, 2013.

\bibitem{1dSfm_11}
Venu~Madhav Govindu.
\newblock Combining two-view constraints for motion estimation.
\newblock In {\em Computer Vision and Pattern Recognition, 2001. CVPR 2001.
  Proceedings of the 2001 IEEE Computer Society Conference on}, volume~2, pages
  II--218. IEEE, 2001.

\bibitem{AMIT_13}
Venu~Madhav Govindu.
\newblock Lie-algebraic averaging for globally consistent motion estimation.
\newblock In {\em Computer Vision and Pattern Recognition, 2004. CVPR 2004.
  Proceedings of the 2004 IEEE Computer Society Conference on}, volume~1, pages
  I--684. IEEE, 2004.

\bibitem{1dSfm_13}
Richard Hartley, Khurrum Aftab, and Jochen Trumpf.
\newblock L1 rotation averaging using the weiszfeld algorithm.
\newblock In {\em Computer Vision and Pattern Recognition (CVPR), 2011 IEEE
  Conference on}, pages 3041--3048. IEEE, 2011.

\bibitem{1dSfm_14}
Nianjuan Jiang, Zhaopeng Cui, and Ping Tan.
\newblock A global linear method for camera pose registration.
\newblock In {\em Computer Vision (ICCV), 2013 IEEE International Conference
  on}, pages 481--488. IEEE, 2013.

\bibitem{1dSfm_16}
Fredrik Kahl.
\newblock Multiple view geometry and the $l^\infty$-norm.
\newblock In {\em Computer Vision, 2005. ICCV 2005. Tenth IEEE International
  Conference on}, volume~2, pages 1002--1009. IEEE, 2005.

\bibitem{AMIT_19}
Fredrik Kahl and Richard Hartley.
\newblock Multiple-view geometry under the $l_\infty$-norm.
\newblock {\em Pattern Analysis and Machine Intelligence, IEEE Transactions
  on}, 30(9):1603--1617, 2008.

\bibitem{LaurentMassart}
B.~Laurent and P.~Massart.
\newblock Adaptive estimation of a quadratic functional by model selection.
\newblock {\em Ann. Statist.}, 28(5):1302--1338, 10 2000.

\bibitem{L2004}
J.~L\"{o}fberg.
\newblock Yalmip : A toolbox for modeling and optimization in {MATLAB}.
\newblock In {\em Proceedings of the CACSD Conference}, Taipei, Taiwan, 2004.

\bibitem{1dSfm_17}
Daniel Martinec and Tomas Pajdla.
\newblock Robust rotation and translation estimation in multiview
  reconstruction.
\newblock In {\em Computer Vision and Pattern Recognition, 2007. CVPR'07. IEEE
  Conference on}, pages 1--8. IEEE, 2007.

\bibitem{1dSfm_18}
Pierre Moulon, Pascal Monasse, and Renaud Marlet.
\newblock Global fusion of relative motions for robust, accurate and scalable
  structure from motion.
\newblock In {\em Computer Vision (ICCV), 2013 IEEE International Conference
  on}, pages 3248--3255. IEEE, 2013.

\bibitem{AMIT}
Onur \"{O}zye\c{s}\.{i}l and Amit Singer.
\newblock Robust camera location estimation by convex programming.
\newblock {\em CoRR}, abs/1412.0165, 2014.

\bibitem{AMIT_25}
Onur \"{O}zye\c{s}\.{i}l, Amit Singer, and Ronen Basri.
\newblock Camera motion estimation by convex programming.
\newblock {\em CoRR}, abs/1312.5047, 2013.

\bibitem{AMIT_27}
Kristy Sim and Richard Hartley.
\newblock Recovering camera motion using $l^\infty$ minimization.
\newblock In {\em Computer Vision and Pattern Recognition, 2006 IEEE Computer
  Society Conference on}, volume~1, pages 1230--1237. IEEE, 2006.

\bibitem{1dSfm_19}
Sudipta~N Sinha, Drew Steedly, and Richard Szeliski.
\newblock A multi-stage linear approach to structure from motion.
\newblock In {\em Trends and Topics in Computer Vision}, pages 267--281.
  Springer, 2012.

\bibitem{TTT1999}
K.~C. Toh, M.J. Todd, and R.~H. Tutuncu.
\newblock Sdpt3 - a matlab software package for semidefinite programming.
\newblock {\em Optimization Methods and Software}, 11:545--581, 1998.

\bibitem{1dSfm_22}
Bill Triggs, Philip~F McLauchlan, Richard~I Hartley, and Andrew~W Fitzgibbon.
\newblock Bundle adjustmentÑa modern synthesis.
\newblock In {\em Vision algorithms: theory and practice}, pages 298--372.
  Springer, 2000.

\bibitem{AMIT_32}
Roberto Tron and Ren{\'e} Vidal.
\newblock Distributed image-based 3-d localization of camera sensor networks.
\newblock In {\em Decision and Control, 2009 held jointly with the 2009 28th
  Chinese Control Conference. CDC/CCC 2009. Proceedings of the 48th IEEE
  Conference on}, pages 901--908. IEEE, 2009.

\bibitem{TTT2003}
R.H. Tutuncu, K.C. Toh, and M.J. Todd.
\newblock Solving semidefinite-quadratic-linear programs using sdpt3.
\newblock {\em Mathematical Programming Ser. B}, 95:189--217, 2003.

\bibitem{Vershynin}
R.~Vershynin.
\newblock Introduction to the non-asymptotic analysis of random matrices.
\newblock In Y.C. Eldar and G.~Kutyniok, editors, {\em Compressed Sensing:
  Theory and Applications}. Cambridge University Press, 2012.

\bibitem{1dsfm}
Kyle Wilson and Noah Snavely.
\newblock Robust global translations with 1dsfm.
\newblock In {\em Proceedings of the European Conference on Computer Vision
  ({ECCV})}, 2014.

\end{thebibliography}

\end{document}